\documentclass[conference]{IEEEtran}
\IEEEoverridecommandlockouts
\usepackage{cite}
\usepackage{amsmath,amssymb,amsfonts,amsthm}
\usepackage{algorithmic}
\usepackage{hyperref}
\hypersetup{colorlinks,allcolors=black}
\usepackage{algorithm}
\usepackage{graphicx}
\usepackage{microtype}
\usepackage{subfigure}
\usepackage{textcomp}
\usepackage{xcolor}
\usepackage{url}
\usepackage{float}
\usepackage{caption}
\usepackage{dsfont}
\def\BibTeX{{\rm B\kern-.05em{\sc i\kern-.025em b}\kern-.08em
    T\kern-.1667em\lower.7ex\hbox{E}\kern-.125emX}}

\allowdisplaybreaks
\newtheorem{lemma}{Lemma}
\newtheorem{thm}{Theorem}
\newtheorem{defn}{Definition}[]

\newtheorem{remark}{Remark}
\newtheorem{assumption}{Assumption}
\usepackage{dblfloatfix}
\begin{document}

\title{Learning-based Scheduling for Information\\ Accuracy and Freshness in Wireless Networks
\author{Hitesh Gudwani \\ \textit{Industrial Engineering and Operations Research Department} \\ \textit{Indian Institute of Technology Bombay} \\ hitesh\_ieor@iitb.ac.in}
}

\maketitle

\begin{abstract}
We consider a system of multiple sources, a single communication channel, and a single monitoring station. Each source measures a time-varying quantity with varying levels of accuracy and one of them sends its update to the monitoring station via the channel. The probability of success of each attempted communication is a function of the source scheduled for transmitting its update. Both the probability of correct measurement and the probability of successful transmission of all the sources are unknown to the scheduler. The metric of interest is the reward received by the system which depends on the accuracy of the last update received by the destination and the Age-of-Information (AoI) of the system. We model our scheduling problem as a variant of the multi-arm bandit problem with sources as different arms. We compare the performance of all $4$ standard bandit policies, namely, ETC, $\epsilon$-greedy, UCB, and TS suitably adjusted to our system model via simulations. In addition, we provide analytical guarantees of $2$ of these policies, ETC, and $\epsilon$-greedy. Finally, we characterize the lower bound on the cumulative regret achievable by any policy.
\end{abstract}


\section{Introduction}
\label{sec:introduction}
This study is motivated by IoT-type devices that require real-time and accurate updates from the sensors to give a good performance. IoT finds applications in many areas such as smart cars, smartphones, healthcare, etc. It turns out that accurate and timely updates from the sensor about the underlying time-varying processes to the destination increase the performance of the system.

We consider a system of multiple sources, a single channel, and a single monitoring station/destination, similar to that in \cite{li2021efficient}, where all the sources measure a common phenomenon and generate an update of their measurement every time slot in order to transmit it to the monitoring station over the communication channel. In a time slot, only $1$ source can transmit its update over the channel. Each source is of a different quality, which means, each source can measure the phenomenon accurately with some probability. In addition, each source while using the channel can send its update over the channel with a successful transmission probability that can vary across sources. It may be possible that the best quality source has the least probability of successful transmission over the channel and vice versa. Both the probability of correct measurement and the probability of successful transmission of all the sources are unknown to the scheduler which requires gradual learning from its local observations. So, it is a sequential decision-making problem where the scheduler makes a decision in every time slot regarding which source should send its update over the channel.

One example of this scenario can be thought of as the sources being cameras, that are positioned at different locations and at different angles, capturing the image of a moving object in every time slot in order to send their images to a monitoring station. All the cameras may capture a different quality image of the object due to various reasons like low resolution or inaccurate positioning of the camera. In addition, some cameras may have better transmission capability for their updates over the channel as compared to others. The scheduler needs to schedule the updates from different cameras such that the destination is updated with an accurate image of the object at all times.

Age-of-Information (AoI) is a popular metric used to capture the freshness of the information received at the destination. It is formally defined as the time elapsed since the destination received the latest update. In our setup, the reward at the destination in every time slot depends on the accuracy of the last update received at the destination depreciated by a factor that depends on the AoI of the system as seen from the destination at that time slot. Hence, the more the AoI of the system at a time slot, the more the depreciation of the last update received. We model this setup as a multi-armed bandit problem where sources act as different arms and our goal is to minimize the cumulative regret of the system, defined as the difference in the cumulative expected reward under the optimal and proposed policy.

\subsection{Our Contributions}
The main contributions of this work are as follows:
\begin{enumerate}
\item[--] We characterize the optimal source that would maximize the cumulative reward received by the destination should we schedule that source throughout the run of the policy.
\item[--] We design ETC, $\epsilon$-greedy, UCB, TS policies for our setting. We provide analytical guarantees for the ETC and $\epsilon$-greedy policy. In our setting, the reward received across time slots is not i.i.d. and depends on the reward received in the previous time slots. Hence, standard proof techniques cannot be used here and a novel proof approach is required.
\item[--] We compare the performance of all the $4$ policies stated above via simulations. Our results show that TS outperforms all the other policies followed by the $\epsilon$-greedy policy.
\end{enumerate}

\subsection{Related Work}
In this section, we discuss the prior work most relevant to our setting.

Studies in various applications considering AoI as metric have been conducted such as AoI optimization (\cite{yates2015lazy}, \cite{altman2019forever}), AoI in energy harvesting (\cite{kaul2012real, sun2017update, dong2020energy}), AoI in vehicular networks (\cite{kaul2011minimizing, choudhury2020experimental}), online sampling for remote estimation (\cite{nar2014sampling, ornee2019sampling}). The ones closer to our work are the scheduling problems with the objective of minimizing the AoI of the system and are studied in \cite{sombabu2018age, tripathi2017age, tripathi2019whittle, jhunjhunwala2018age, hsu2019scheduling, kadota2018optimizing}. However, these studies have been conducted for an infinite time horizon considering the channel statistics to be known whereas we are analyzing the performance of our system for a finite time horizon considering system parameters to be unknown to the scheduler. \cite{shreedhar2018acp, yates2018age, yates2020age, talak2017minimizing, bedewy2019age} studies AoI-based scheduling problems in multi-hop networks. The main distinction between this and our work is that we consider a setting where channel parameters are unknown and need to be learned.

Closest to our work, \cite{li2021efficient} considers the system that has multiple sources of different degrees of importance where more than $1$ source can transmit its update in a time slot to the destination. Hence, the scheduler needs to decide on a set of sources in every time slot that would transmit their updates to the destination. While multiple sources sending their updates simultaneously in a time slot is not possible in all cases, in our work, we consider a single channel due to which only $1$ source can transmit its update in a time slot. Also, in \cite{li2021efficient}, while authors consider a dual objective function of minimizing regret and age, we merge these two by the way we define the reward received by the system in every time slot which includes the accuracy of the update last received by the destination and also the age of the system as seen from the destination at that time slot. This necessitates modifications in the policy as well as novel proof methodology.

Other body of work includes \cite{fatale2021regret, deshpande2021age, krishnasamy2016regret} where AoI has been optimized considering different system models. These settings are significantly different and cannot be readily adapted to our setting but our proofs are motivated by various proof steps used in these papers.

\section{Problem Setup}
\label{sec:setup}
\subsection{System Model}
\label{subsec:our system}

We consider a system of $K$ sources, $1$ communication channel, and a monitoring station. Time is divided into slots.

Each source measures a common phenomenon with a certain level of accuracy in every time slot and one of them transmits its updates to the destination via a communication  channel with some probability. For a source $k$, the probability of successfully transmitting over the channel is given by $p_k$, and the probability of measuring the phenomenon accurately is given by $q_k$, $\forall~ k \in [1, K]$. It follows that the channel is ON-OFF type which is i.i.d across time slots and independent across sources i.e., irrespective of the time slot the channel is ON for the source $k$ with probability $p_k$ and OFF otherwise, independent of all other events. Also, the update from the source $k$ is GOOD-BAD type which is again i.i.d. across time slots and independent across sources i.e., the update transmitted, if it has, is GOOD or a correct measurement of the phenomenon with probability $q_k$ and BAD otherwise. We assume in this setting that $p_k$ and $q_k$ are unknown to the scheduler.

The notation used in describing the problem setup is given in Table \ref{table: problem setup}.

\begin{table}
\centering
\caption{Notation used in the problem setup}
\label{table: problem setup}
\begin{tabular}{|c|c|}
\hline
\textbf{Symbol} & \textbf{Description} \\ \hline \hline
$p_k$ & Probability of successful transmission for source $k$ \\ \hline
$q_k$ & Probability of accurate measurement by source $k$ \\ \hline
$K$ & Total number of sources \\ \hline
$d$ & Depreciating factor $\in (0,1)$ \\ \hline
\end{tabular}
\end{table}
\subsection{Performance Metric}

The Age-of-Information (AoI) of the system at the monitoring station is defined as follows:
\begin{defn}[Age-of-Information (AoI)]
Let $a(t)$ be the AoI of the system at the monitoring station in time slot $t$ and $u(t)$ denote the time-slot in which the monitoring station received the latest update before time slot $t$ irrespective of the source that transmitted it. Then, $$a(t) = t - u(t)$$.
This implies,
\begin{equation*}
    a(t) =
    \begin{cases}
    1 & \text{if the transmission in time}\\
     & \text{slot $t-1$ succeeds}\\
    a(t-1) + 1 & \text{otherwise}.
    \end{cases}
\end{equation*}   
\end{defn}
Let us assume $a(t)$ to be the AoI of the system at the monitoring station in time slot $t$ under a given scheduling policy $\mathcal{P}$. This implies that the last update received by the monitoring station before time $t$ is $Q(t-a(t)) ~ (\in \{0,1\})$ from a source. Let us also assume a constant, $d$, called the depreciating factor of the system where $d\in (0,1)$ and is known to the scheduler. We define the reward received by the system at the monitoring station in time slot $t$ under the policy $\mathcal{P}$ as follows:
\begin{defn}[Reward under policy $\mathcal{P}$]
Reward received by the system at the monitoring station under policy $\mathcal{P}$ in time-slot $t$ is denoted by $r^\mathcal{P}(t)$ and
\begin{equation*}
    r^\mathcal{P}(t)=Q(t-a(t))d^{(a(t)-1)}
\end{equation*}
where d is a depreciating factor; $d\in (0,1)$.\\

This implies,
\begin{equation*}
    r^\mathcal{P}(t)=
    \begin{cases}
    Q(t-1) & \text{if the transmission in time}\\
     & \text{slot $t-1$ succeeds}\\
    r^\mathcal{P}(t-1)d & \text{otherwise}.
    \end{cases}
\end{equation*}
\end{defn}
Let $r^\mathcal{P}(t)$ be the reward received at the destination in time-slot $t$ under a given scheduling policy $\mathcal{P}$ and $r^*(t)$ be the corresponding reward under the oracle policy, which we define in section \ref{sec:oracle}. We define cumulative regret under policy $\mathcal{P}$ as the sum of the difference in the expected rewards under the oracle policy and the policy $\mathcal{P}$ in time-slots 1 to $T$. Formally, cumulative reward under policy $\mathcal{P}$ is defined as follows:
\begin{defn}[Cumulative Regret]
\label{def:regret}
Cumulative regret under policy $\mathcal{P}$ from time-slot $1$ to $T$ is denoted by $\mathcal{R}_\mathcal{P}(T)$ where
\begin{equation}
\label{eq:regret definition}
    \mathcal{R}_\mathcal{P}(T)=\sum_{t=1}^T \mathbb{E}[r^*(t)-r^\mathcal{P}(t)].
\end{equation}
\end{defn}

The following assumption is made on the initial state of the system for concreteness and technical convenience.

\begin{assumption}[Initial conditions for a candidate policy]
The system starts operating in time slot $t=-\infty$ and a source sends its update to the monitoring station via the communication channel in every time slot up to $t=0$. The candidate policy starts executing at $t=1$. The AoI of the system is initialized to $0$ at $t = 0$, i.e., a(0) = 0 and the observations in time slots $t \leq 0$ are not used in scheduling the decisions in time slots $t \geq 1$.
\end{assumption}

\subsection{Goal}
Our goal in this paper is to design a scheduling policy/algorithm to determine the source which will transmit its update through the communication channel in every time slot based on past observations and their outcomes in order to minimize the cumulative regret over the time horizon, T (Definition \ref{def:regret}).

\subsection{Oracle Policy}
\label{sec:oracle}
Under the oracle policy, the scheduler knows the value of $p_k$ and $q_k$ for all the sources $k\in [1,K]$. As a result, it schedules the optimal source, say $k^*$ (see Theorem \ref{thm1} for characterization of the optimal source), in every time slot $t$ to transmit its update over the communication channel.

The following assumption is made for oracle policy for concreteness and technical convenience.

\begin{assumption}[Initial conditions for oracle policy]
 The system starts at $t=-\infty$ and the scheduler schedules the optimal source, $k^*$, in every time slot $t$. 
\end{assumption}

\color{black}

\section{Main Results and Discussion}
\label{sec:ST algorithm}

In this section, we state and discuss our main results.

\subsection{Characterization of Oracle Policy}
We characterize the optimal source for our setting. The oracle policy uses the optimal source in all time slots.
\begin{thm}
\label{thm1}
Given a problem instance $(p_k, q_k, d, K)$, the oracle policy uses the optimal source, $k^*$, given as
\begin{equation*}
    k^* = \arg \max_{1 \leq k \leq K}\Bigg(\frac{p_kq_k}{1-d(1-p_k)}\Bigg)
\end{equation*}
\end{thm}
The proof of this theorem is given in the appendix.
\subsection{Scheduling Policies}
\label{sec: scheduling policies}
We propose $4$ standard bandit policies adopted for our setting. The key idea behind the implementation of all these policies is to use the result of theorem \ref{thm1} to steer the system towards using the optimal source. We first present a description of the policies and then provide a formal definition of each of them. Let us define 2 notations that are used in all the policies:\\
$P_k = \mathds{1}\{\text{successful transmission by source $k$}\}$\\
$PQ_k = \mathds{1}\{\text{successful transmission and accurate measurement}\\
~~~~~~~~~~~~~~~\text{by source $k$}\}$\\

\subsubsection{Explore then Commit (ETC) Policy} This policy runs in two phases, the explore phase and then the commit phase. In the explore phase, all the sources transmit their updates to the monitoring station in a round-robin manner through the communication channel. The goal of this phase is to let the monitoring station observe a sufficient number of samples of
\begin{enumerate}
\item[(i)] $PQ_k~\forall~1 \leq k \leq K$.
\item[(ii)] $P_k ~ \forall ~ 1\leq k\leq K$
\end{enumerate}
so that the monitoring station can use these local observations to maintain an empirical estimate of $p_kq_k$ and $p_k$ for each source, $k$, in order to get to the optimal source, $\hat{k}$, with sufficiently high probability, by calculating the expression given in Theorem \ref{thm1}. The duration of the explore phase, denoted by $T_E$, is an input to the algorithm. Intuitively, the value of $T_E$ should be large enough to ensure that the monitoring station correctly gets the empirical estimate of $p_kq_k$ and $p_k$ for each source with high probability using its local observations so that it gets the correct $\hat{k}$ from Theorem \ref{thm1} with high probability.

From time slot $T_E+1$ onwards, the algorithm enters the commit phase. At the beginning of the commit phase, the monitoring station calculates the $\hat{k}$ value from theorem \ref{thm1} by using the empirical estimates of $p_kq_k$ and $p_k$ for each source which it calculates from the samples obtained of (i) and (ii) respectively till the end of the explore phase and then continues to use that source till the end of the algorithm. 

Refer to Algorithm \ref{Algo1} for the formal definition of the ETC policy.

\subsubsection{$\epsilon$ - greedy Policy} In every time slot $t$, the monitoring station selects a source uniformly at random with probability $\epsilon_t$, which is a function of $t$, and selects an empirically best source, $\hat{k}$, otherwise, which it gets from Theorem \ref{thm1} after calculating the empirical estimates of $p_kq_k$ and $p_k$ from its local observations received till that time slot. The former is called Exploration and the latter, Exploitation. Additionally, in the explore time slot, the monitoring station updates the empirical estimate of $p_kq_k$ and $p_k$ of the source, $k$, selected in that time slot after getting its local observations but in the exploit time slot, it does not do so.

The goal of this algorithm is to initially do more exploration to get the correct empirical estimates of $p_kq_k$ and $p_k$ of all the sources and gradually shift to more exploitation after it gets to know the optimal source with high probability. 

Refer to Algorithm \ref{algo: epsilon greedy} for the formal definition of $\epsilon$-greedy policy.

\subsubsection{Upper Confidence Bound (UCB) Policy}
After scheduling each source $1$ time, in every time slot, the monitoring station calculates the upper confidence bound ($UCB_k(t)$) for every source $k$ and selects the source with the maximum $UCB_k(t)$. It is defined as follows:\\
\begin{equation*}
    UCB_k(t) = \frac{UCB({PQ_k}(t))}{1-d(1-LCB(P_k(t)))}
\end{equation*}
where,
\begin{align*}
    UCB({PQ_k}(t)) &= \hat{PQ_k}(t) + \sqrt{\frac{2\log t}{u_k(t)}}\\
    LCB({P_k}(t)) &= \max\Bigg\{0,\hat{P_k}(t) - \sqrt{\frac{2\log t}{u_k(t)}}\Bigg\}\\
\end{align*}
Here, $\hat{PQ_k}(t)$ and $\hat{P_k}(t)$ are the empirical estimates of $p_kq_k$ and $p_k$ respectively for source $k$ till time slot $t$ by using the local observations received by the monitoring station and $u_k(t)$ is the number of times source $k$ has been scheduled for transmission till time slot $t$.

Refer to Algorithm \ref{algo: UCB} for the formal definition of upper confidence bound (UCB) policy.
\subsubsection{Thompson Sampling (TS) Policy} In every time slot $t$, the monitoring station selects a source $k(t)$ such that\\
\begin{equation*}
    k(t) = \arg \max_{1 \leq k \leq K}\Bigg(\frac{x_k(t)}{1-d(1-y_k(t))}\Bigg)
\end{equation*}
where,
\begin{align*}
    x_k(t) &\sim Beta(S(PQ_k(t))+1,F(PQ_k(t))+1)\\
    y_k(t) &\sim Beta(S(P_k(t))+1,F(P_k(t))+1)\\
\end{align*}
Here, $S(PQ_k(t))$ and $F(PQ_k(t))$ are the number of successes (meaning $1's$) and failures (meaning $0's$) respectively of $PQ_k$ for source $k$ till time slot $t$. Likewise, $S(P_k(t))$ and $F(P_k(t))$ are the number of successes (meaning $1's$) and failures (meaning $0's$) respectively of $P_k$ for source $k$ till time slot $t$.

Refer to Algorithm \ref{algo: thompson Sampling} for the formal definition of Thompson Sampling (TS) policy.
\color{black}

\begin{algorithm}[!t]

	\caption{Explore-then-Commit (ETC)} \label{Algo1}
	\begin{algorithmic}[1]
		\STATE	Input: {$p_k, q_k ~\forall ~K$, $d, T, T_E (\text{duration of explore phase})$ }
		\STATE $\text{Initialize:}~PQ_k = P_k = 0$ ($PQ$ \& $P$ sample for source $k$)\\~~~~~~~~~~~ $\hat{PQ_k} = \hat{P_k} = 0$ ($p_kq_k$ \& $p_k$ estimate - source $k$)\\ ~~~~~~~~~~~~$\hat{\mu_k} = 0 ~\forall ~ K$ ($\mu_k$ estimate for source $k$)\\~~~~~~~~~~~ $T_k = 0 ~\forall ~K$ (\# source $k$ is scheduled)\\~~~~~~~~~~~~$a(0) = 0 ~\forall K$ (age at time slot $0$)\\~~~~~~~~~~~~$r^{\mathcal{ETC}}(0) = 0$ (reward under $ETC$ at time $0$)
        \WHILE{$1 \leq t \leq T_E$}
        \FOR{$k=1,2, \hdots, K$}
        \STATE Schedule update from source $k(t) = k$
        \STATE Receive transmission sample $P_{k(t)} \sim \text{Ber}(p_{k(t)})$
        \IF{($P_{k(t)} == 0$)}
        \STATE $a(t) = a(t-1) +1$
        \STATE $r^{\mathcal{ETC}}(t) = Q_{k(t-1)} d^{a(t)-1}$
        \STATE $T_{k(t)} = T_{k(t)} + 1$
        \ELSE
        \STATE Receive update sample $Q_{k(t)} \sim \text{Ber}(q_{k(t)})$
        \STATE $a(t) = 1$
        \STATE $r^{\mathcal{ETC}}(t) = Q_{k(t)}$
        \STATE $PQ_{k(t)} = PQ_{k(t)} + P_{k(t)}Q_{k(t)}$
        \STATE $P_{k(t)} = P_{k(t)} + 1$
        \STATE $T_{k(t)} = T_{k(t)} + 1$
        \ENDIF
        \STATE $t=t+1$
        \ENDFOR
        \ENDWHILE
        \FOR{$k=1,2, \hdots, K$}
        \STATE $\hat{PQ_k} = \frac{PQ_k}{T_k}$
        \STATE $\hat{P_k} = \frac{P_k}{T_k}$
        \STATE $\hat{\mu_k} = \frac{\hat{PQ_k}}{1-d(1-\hat{P_k})}$
        \ENDFOR
        \FOR{$t=T_E+1, 2, \hdots, T$}
        \STATE schedule update from $k(t) = \arg \max\limits_{1 \leq k \leq K}\hat{\mu_k} $
        \STATE Repeat steps $6$ to $18$
        \ENDFOR
        \STATE $r^{oracle} = \text{Oracle}(p_k, q_k, d, T)$
        \FOR{$t=1, 2, \hdots, T$}
        \STATE $\mathcal{R}_\mathcal{ETC}(t) = r^{oracle}(t) - r^{\mathcal{ETC}}(t)$ (Regret at time $t$)
        \ENDFOR
        \FOR{$t=1, 2, \hdots, T$}
        \STATE $\mathcal{R}_\mathcal{ETC}(T) = \mathcal{R}_\mathcal{ETC}(T) + \mathcal{R}_{\mathcal{ETC}}(t)$ (Cum. Regret till $T$)
        \ENDFOR
        \STATE Output: {$\mathcal{R}_{\mathcal{ETC}}(T)$}
	\end{algorithmic}
\end{algorithm}

\begin{algorithm}[!h]
	\caption{$\epsilon$- greedy}
        \label{algo: epsilon greedy}
	\begin{algorithmic}[1]
		\STATE	Input: {$p_k, q_k ~\forall ~K$, $d, T, \epsilon_t$ }
		\STATE $\text{Initialize:}~PQ_k = P_k = 0$ ($PQ$ \& $P$ sample for source $k$)\\~~~~~~~~~~~ $\hat{PQ_k} = \hat{P_k} = 0$ ($p_kq_k$ \& $p_k$ estimate - source $k$)\\ ~~~~~~~~~~~~$\hat{\mu_k} = 0 ~\forall ~ K$ ($\mu_k$ estimate for source $k$)\\~~~~~~~~~~~ $T_k = 0 ~\forall ~K$ (\# source $k$ is scheduled)\\~~~~~~~~~~~~$a(0) = 0 ~\forall K$ (age at time slot $0$)\\~~~~~~~~~~~~$r^{\epsilon}(0) = 0$ (reward under $\epsilon$-greedy at time $0$)
        \FOR{$t=1, 2, \hdots, T$}
        \STATE Receive $X_t \sim \text{Ber}(\epsilon_t)$
        \IF{($X_t == 0$)}
        \STATE Schedule update from source $k(t)$ chosen uniformly at random
        from $\{1, 2, \hdots, K\}$
        \STATE Repeat steps $6$ to $18$ in Algorithm \ref{Algo1} replacing $r^{\mathcal{ETC}}(t)$ with $r^{\epsilon}(t)$
        \STATE $\hat{PQ}_{k(t)} = \frac{PQ_{k(t)}}{T_{k(t)}}$
        \STATE $\hat{P}_{k(t)} = \frac{P_{k(t)}}{T_{k(t)}}$
        \STATE $\hat{\mu}_{k(t)} = \frac{\hat{PQ}_{k(t)}}{1-d(1-\hat{P}_{k(t)})}$
        \ELSE
        \STATE Schedule update from $k(t) = \arg \max\limits_{1 \leq k \leq K} \hat{\mu}_{k(t)}$
        \STATE Repeat steps $6$ to $18$ in Algorithm \ref{Algo1} replacing $r^{\mathcal{ETC}}(t)$ with $r^{\epsilon}(t)$
        \ENDIF
        \ENDFOR
        \STATE $r^{oracle} = \text{Oracle}(p_k, q_k, d, T)$
        \FOR{$t=1, 2, \hdots, T$}
        \STATE $\mathcal{R}_{\epsilon}(t) = r^{oracle}(t) - r^{\epsilon}(t)$ (Regret at time $t$)
        \ENDFOR
        \FOR{$t=1, 2, \hdots, T$}
        \STATE $\mathcal{R}_{\epsilon}(T) = \mathcal{R}_{\epsilon}(T) + \mathcal{R}_{\epsilon}(t)$ (Cum. Regret till $T$)
        \ENDFOR
        \STATE Output: {$\mathcal{R}_{\epsilon}(T)$}
	\end{algorithmic}
\end{algorithm}
\color{black}

\begin{algorithm}[!h]
	\caption{Upper Confidence Bound (UCB)}
        \label{algo: UCB}
	\begin{algorithmic}[1]
		\STATE Input: {$p_k, q_k ~\forall ~K$, $d, T$} 
        \STATE $\text{Initialize:}~PQ_k = P_k = 0$ ($PQ$ \& $P$ sample for source $k$)\\~~~~~~~~~~~ $\hat{PQ_k} = \hat{P_k} = 0$ ($p_kq_k$ \& $p_k$ estimate - source $k$)\\ ~~~~~~~~~~~~$UCB_k = 0 ~\forall ~ K$ ($UCB$ for source $k$)\\~~~~~~~~~~~ $T_k = 0 ~\forall ~K$ (\# source $k$ is scheduled)\\~~~~~~~~~~~~$a(0) = 0 ~\forall K$ (age at time slot $0$)\\~~~~~~~~~~~~$r^\mathcal{UCB}(0) = 0$ (reward under $UCB$ at time $0$)
		\FOR{$t=1, 2, \hdots, K$}
		  \STATE Schedule update from source $k(t) = t$
            \STATE Repeat steps $6$ to $18$ in Algorithm \ref{Algo1} replacing $r^\mathcal{ETC}(t)$ with $r^\mathcal{UCB}(t)$
            \STATE Repeat steps $8$ and $10$ in Algorithm \ref{algo: epsilon greedy} to get $\hat{PQ}_{k(t)}$and $\hat{P}_{k(t)}$
        \ENDFOR
        \FOR{$t=K+1, K+2, \hdots, T$}
            \FOR{$k=1, 2, \hdots, K$}
                \STATE $UCB(PQ_k) = \hat{PQ_k} + \sqrt{\frac{2\log t}{T_k}}$
                \STATE $LCB(P_k) = \max\Big\{0,\hat{P_k} - \sqrt{\frac{2\log t}{T_k}}\Big\}$
                \STATE $UCB_k = \frac{UCB({PQ_k})}{1-d(1-LCB({P_k}))}$
            \ENDFOR
            \STATE Schedule $k(t) = \arg \max\limits_{1 \leq k \leq K}UCB_k$
            \STATE Repeat steps $6$ to $18$ in Algorithm \ref{Algo1} replacing $r^\mathcal{ETC}(t)$ with $r^\mathcal{UCB}(t)$
            \STATE Repeat steps $8$ and $10$ in Algorithm \ref{algo: epsilon greedy} to get $\hat{PQ}_{k(t)}$and $\hat{P}_{k(t)}$
        \ENDFOR
        \STATE $r^{oracle} = \text{Oracle}(p_k, q_k, d, T)$
        \FOR{$t=1, 2, \hdots, T$}
        \STATE $\mathcal{R}_\mathcal{UCB}(t) = r^{oracle}(t) - r^\mathcal{UCB}(t)$ (Regret at time $t$)
        \ENDFOR
        \FOR{$t=1, 2, \hdots, T$}
        \STATE $\mathcal{R}_\mathcal{UCB}(T) = \mathcal{R}_\mathcal{UCB}(T) + \mathcal{R}_\mathcal{UCB}(t)$ (Cum. Regret till $T$)
        \ENDFOR
        \STATE Output: {$\mathcal{R}_\mathcal{UCB}(T)$}
	\end{algorithmic}
\end{algorithm}

\begin{algorithm}[!t]
	\caption{Thompson Sampling}
        \label{algo: thompson Sampling}
	\begin{algorithmic}[1]
		\STATE Input: {$p_k, q_k ~\forall ~K$, $d, T$} 
        \STATE $\text{Initialize:}~P_k = Q_k = 0$ ($P$ \& $Q$ sample for source $k$)\\~~~~~~~~~~~ $S(PQ_k) = S(P_k) = 0$ (\# of $1$'s of $PQ_k$ $\&$ $P_k$)\\ ~~~~~~~~~~~~$F(PQ_k) = F(P_k) = 0$ (\# of $0$'s of $PQ_k$ $\&$ $P_k$)\\~~~~~~~~~~~~$X_k = Y_k = \mu_k = 0$\\~~~~~~~~~~~~$a(0) = 0 ~\forall K$ (age at time slot $0$)\\~~~~~~~~~~~~$r^\mathcal{TS}(0) = 0$ (reward under $TS$ at time $0$)
        \FOR{$t=1, 2, \hdots, T$}
            \FOR{$k=1, 2, \hdots, K$}
                \STATE Receive $X_{k} \sim \text{Beta}(S(PQ_k)+1, F(PQ_k)+1)$
                \STATE Receive $Y_k \sim \text{Beta}(S(P_k)+1, F(P_k)+1)$
                \STATE $\mu_k = \frac{X_k}{1-d(1-Y_k)}$
            \ENDFOR
            \STATE Schedule update from $k(t) = \arg \max\limits_{1 \leq k \leq K}\mu_k$
            \STATE Receive transmission sample $P_{k(t)} \sim \text{Ber}(p_{k(t)})$
            \IF{($P_{k(t)} == 0$)}
                \STATE $a(t) = a(t-1) +1$
                \STATE $r^{\mathcal{TS}}(t) = Q_{k(t-1)} d^{a(t)-1}$
                \STATE $F(PQ_{k(t)}) = F(PQ_{k(t)})+1$
                \STATE $F(P_{k(t)}) = F(P_{k(t)})+1$
            \ELSE
                \STATE Receive update sample $Q_{k(t)} \sim \text{Ber}(q_{k(t)})$
                \STATE $a(t) = 1$
                \STATE $r^{\mathcal{TS}}(t) = Q_{k(t)}$
                \STATE $S(P_{k(t)}) = S(P_{k(t)})+1$
                \IF{($Q_{k(t)} == 0$)}
                    \STATE $F(PQ_{k(t)}) = F(PQ_{k(t)})+1$
                \ELSE
                \STATE $S(PQ_{k(t)}) = S(PQ_{k(t)})+1$
                \ENDIF
            \ENDIF
        \ENDFOR
        \STATE $r^{oracle} = \text{Oracle}(p_k, q_k, d, T)$
        \FOR{$t=1, 2, \hdots, T$}
        \STATE $\mathcal{R}_\mathcal{TS}(t) = r^{oracle}(t) - r^\mathcal{TS}(t)$ (Regret at time $t$)
        \ENDFOR
	\end{algorithmic}
\end{algorithm}

\begin{algorithm}[!h]
	\caption*{\textbf{Subroutine:} Oracle}
	\begin{algorithmic}[1]
		\STATE Input: {$p_k, q_k ~\forall ~K, d, T$ } 
		\FOR{$k=1, 2, \hdots, K$}
        \STATE $\mu_k = \frac{p_kq_k}{1-d(1-p_k)}$
        \ENDFOR
        \FOR{$t=1, 2, \hdots, T$}
        \STATE Schedule update from $k(t) = \arg \max\limits_{1 \leq k \leq K}\mu_k $
        \STATE Repeat steps $6$ to $18$ in Algorithm \ref{Algo1} replacing $r^{\mathcal{ETC}}(t)$ with $r^{oracle}(t)$
        \ENDFOR
        \STATE Output: {$r^{oracle}$}	
	\end{algorithmic}
\end{algorithm}


\subsection{Analytical Results}
\label{thms}
In this section, we provide our main analytical results. Firstly, we provide performance guarantees for the ETC and $\epsilon$- greedy policies, and then we provide a lower bound on the regret for any $\gamma$ - consistent policy. Our performance guarantees are in the form of fairly complex expressions that depend on all the system parameters. We summarize the results in Table \ref{Table:Order} to highlight the dependence of our regret guarantees for the ETC, $\epsilon$ -greedy, and $\gamma$ - consistent policy on time. The regret bounds on ETC and $\epsilon$ - greedy algorithm holds for $\alpha > 1$ which is a hyper-parameter in both the policies and $c = \max\Big\{\frac{-2}{\ln (d(1-p_{k^*}))}, \frac{4K}{\Delta^2(1-d)^2}\Big\}$ where, $\Delta = \mu_{k^*} - \max\limits_{k\neq k^*}\mu_k$,  $\mu_{k^*} = \frac{p_{k^*}q_{k^*}}{1-d(1-p_{k^*})}$ and $\mu_k = \frac{p_{k}q_{k}}{1-d(1-p_{k})}$.

\begin{table}[h!]
\centering
\caption{Summary of our analytical results}
\begin{tabular}{|c|c|c|}
\hline
 & \textbf{Algorithm} & \textbf{Regret} \\ \hline \hline
1 & Explore-then- Commit $(ETC)$ & $O(K\log T)$ \\ \hline
2 & $\epsilon$-greedy & $O(K^2\log^4T)$ \\ \hline
3 & Any $\gamma$- consistent policy & $\Omega(K\log T)$ \\ \hline
\end{tabular}
\label{Table:Order}
\end{table}

Let $\mathcal{R}_{ETC}(T)$ denote the cumulative regret under the ETC policy from time slot $1$ to $T$ as defined in \eqref{eq:regret definition} (Section \ref{sec:setup}). Our next theorem characterizes the regret of ETC.

\begin{thm}[Performance of ETC policy]
\label{thm: ETC}
Let $p_{k^*}$ and $q_{k^*}$ be the probability of successful transmission and the probability of accurate measurement of the optimal source, $k^*$, as characterized in theorem \ref{thm1}. Then for $\alpha > 1$, $c = \max\Big\{\frac{-2}{\ln (d(1-p_{k^*}))}, \frac{4K}{\Delta^2(1-d)^2}\Big\}$ and $T_E (\text{duration of the explore phase}) = c\ln T$,
\begin{equation*}
    \mathcal{R}_{ETC}(T) \leq \frac{p_{k^*}q_{k^*}}{1-d(1-p_{k^*})}\Bigg[\alpha c\ln T \Bigg(1+\frac{K}{T^4}\Bigg)+\Bigg(\frac{K}{T}+\frac{1}{T^3}\Bigg)\Bigg]
\end{equation*}
\end{thm}

Let $\mathcal{R}_{\epsilon-\text{greedy}}(T)$ denote the cumulative regret under the $\epsilon$- greedy policy from time slot 1 to T as defined in \eqref{eq:regret definition} (Section \ref{sec:setup}). Our next theorem characterizes the regret of $\epsilon$- greedy.
\begin{thm}[Performance of $\epsilon$- greedy]
\label{thm: epsilon greedy}
Let $p_{k^*}$ and $q_{k^*}$ be the probability of successful transmission and the probability of accurate measurement of the optimal source, $k^*$, as characterized in theorem \ref{thm1}. Then for $\alpha > 1$, $c = \max\Big\{\frac{-2}{\ln (d(1-p_{k^*}))},\frac{4K}{\Delta^2(1-d)^2}\Big\}$ and $\epsilon_t = \min\Big\{1,3K\frac{\ln^2 t}{t}\Big\}$,
\begin{align*}
    \mathcal{R}_\epsilon(T)&\leq \frac{p_{k^*}q_{k^*}}{1-d(1-p_{k^*})}\Bigg[\alpha c\ln T + (K-1)\\
    &\quad\times\frac{\frac{c}{K}}{\Big(1-\frac{c}{K}\Big)}\Bigg(\frac{1}{(\alpha-1)c\ln T+1}\Bigg)^{\frac{1-c/K}{c/K}}\\
    &\quad + Kc\ln T\ln^3(T-c\ln T+1)\\
    &\quad + \frac{1}{T} +\frac{c^2\ln T (K-1)}{T^2}\\
    &\quad\times\exp\Bigg({\frac{-\ln^3(\alpha c\ln T-c\ln T+1)}{\frac{c}{K}}}\Bigg)\Bigg]
\end{align*}
\end{thm}

\noindent \textit{Proof Outline for Theorem \ref{thm: ETC} and \ref{thm: epsilon greedy}}: Since the goal of any policy is to select the optimal source, $k^*$, as characterized by Theorem \ref{thm1} in every time slot in order to maximize the cumulative reward obtained over all the time slots, we call the event of selecting the optimal source in a time slot a ``good event". The proof of ETC and $\epsilon$-greedy consists of $2$ main steps. First, we calculate the lower bound of the expected reward conditioned on the fact that the ``good event" will occur in every time slot after the initial learning phase i.e. our policy will only select the optimal source in every time slot after some time. As a second step, we find the lower bound of the probability of occurrence of the ``good event".
\begin{align*}
    \mathbb{E}[r^\mathcal{P}(t)] &\geq \mathbb{E}[r^\mathcal{P}(t)|\text{Occurance of ``good event"}]\\
    &\quad\times\mathbb{P}(\text{Occurance of ``good event"})\\
    \mathcal{R}_\mathcal{P}(T) &\leq \sum_{t=1}^{T} \mathbb{E}[r^*(t) - r^\mathcal{P}(t)]
\end{align*}
The detailed proofs of theorem $2$ and $3$ are given in the appendix. ~~~~~~~~~~~~~~~~~~~~~~~~~~~~~~~~~~~~~~~~~~~~~~~~~~~~~~~~~~~~~~~$\qed$

\begin{remark}
    We thus conclude that cumulative regret scales as $O(K\log T)$ for the $ETC$ policy and scales as $O(K^2\log^4T)$ for the $\epsilon$- greedy policy.
\end{remark} 
Next, we characterize the limit on the performance of any $\gamma$-consistent policy defined as follows:
\begin{defn}[$\gamma$-consistent policies]
Let k(s) denote the index of the source scheduled in time slot $s$ and let $k^*$ be the optimal source as characterized in Theorem \ref{thm1}, i.e. $k^* = \arg \max\limits_{1 \leq k \leq K}\frac{p_kq_k}{1-d(1-p_k)}$. A scheduling policy is said to be a $\gamma$-consistent policy for $\gamma \in (0,1)$, if for any source configuration vector $\boldsymbol{\mu} ~\Big(=\Big\{\mu_k ~:~\mu_k=\frac{p_kq_k}{1-d(1-p_k)}, 1 \leq k \leq K\Big\}$\Big), there exists a constant $C(\boldsymbol{\mu})$ such that
\begin{equation*}
    \mathbb{E}\Bigg[\sum_{s=1}^{t}\mathds{1}\{k(s)=k\}\Bigg] \leq C(\boldsymbol{\mu})t^\gamma ~~ \forall ~~k\neq k^*
\end{equation*}
\end{defn}
\begin{thm}[Lower bound on any $\gamma$- consistent policy]
\label{thm:lower bound}
Given a problem instance $(p_k,q_k,d,K)$, let $\mu_k = \frac{p_kq_k}{1-d(1-p_k)} ~ \forall ~ 1 \leq k \leq K$, $\mu_{k_{\text{min}}} = \min\limits_{k=1:K}\mu_k$, $k^*=\arg \max\limits_{1 \leq k \leq K}\mu_k$ (as defined in Theorem \ref{thm1}), $\mu_{k^*} = \max\limits_{k=1:K}\mu_k = \frac{p_{k^*}q_{k^*}}{1-d(1-p_{k^*})}$, $\Delta = \mu_{k^*} - \max\limits_{k\neq k^*} \mu_k$, $\Delta_p = \min\limits_{k\neq k^*}p_k-p_{k^*}$, $p_\text{min} = \min\limits_{k=1:K}p_k$, $q_{\text{min}} = \min\limits_{k=1:K}q_k$. For any $\gamma$- consistent policy $\mathcal{P}$,
\begin{flalign*}
    \mathcal{R}_\mathcal{P}(T)&\geq \frac{(K-1)D(\mu)\Delta_p}{\Delta}\mu_{k^*}\\
    &\quad\times d((1-\gamma)\log T - \log (4KC))\\
    &\quad +(p_{k^*}q_{k^*} -p_{\text{min}}q_{\text{min}})T
\end{flalign*}
\hspace*{4cm} where, $D(\mu) = \frac{\Delta}{KL\Big(\mu_{k_{\text{min}}},\frac{\mu_{k^*}+1}{2}\Big)}$
\end{thm}
\color{black}

\begin{remark}
    We thus conclude that the cumulative regret of any $\gamma$- consistent policy scales as $\Omega(K\log T)$.
\end{remark}

The proof of theorem $4$ is given in the appendix.

\section{Simulation Results}
\label{sec:simulations}
\begin{figure*}[t]
	\centering     
	\subfigure[]{\label{i1}\includegraphics[width=80mm]{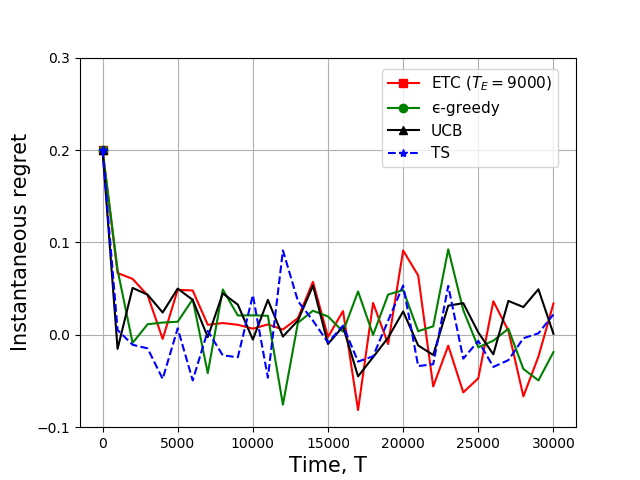}} 
	\subfigure[]{\label{c1}\includegraphics[width=80mm]{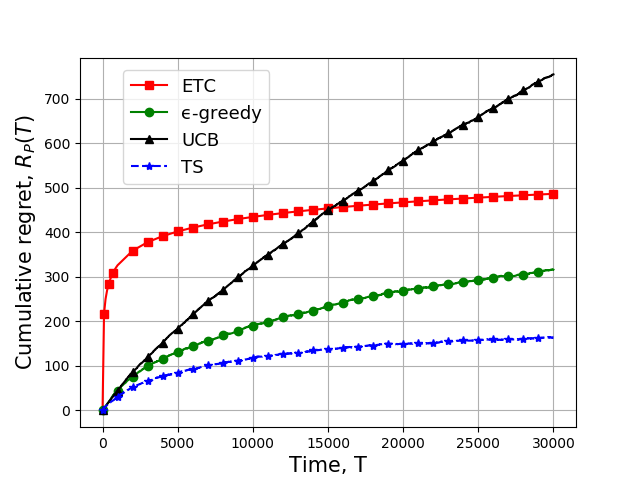}}
	\caption{(a) Instantaneous regret for all the four policies (ETC having $T_E = 9000$) for $K=4$, $p_k = [0.65, 0.7, 0.75, 0.8]$, $q_k = [0.8, 0.75, 0.7, 0.65]$, $d = 0.8$ (b)  Cumulative regret for all the four policies for $K=4$, $p_k = [0.65, 0.7, 0.75, 0.8]$, $q_k = [0.8, 0.75, 0.7, 0.65]$, $d = 0.8$}
	\label{2 plots: K=4}
\end{figure*}

\begin{figure*}
	\centering     
	\subfigure[]{\label{i2}\includegraphics[width=80mm]{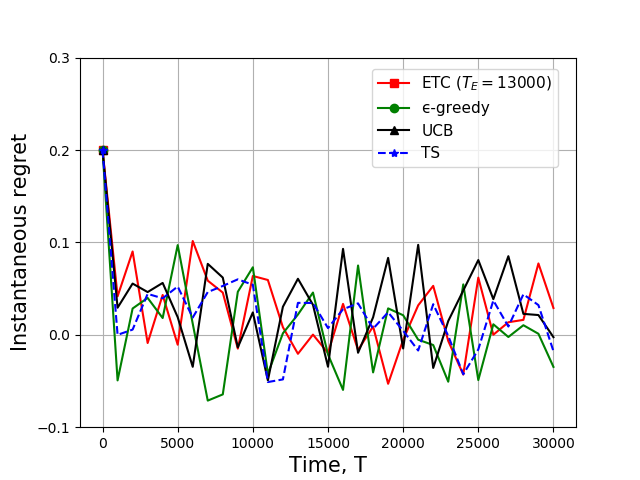}} 
	\subfigure[]{\label{c2}\includegraphics[width=80mm]{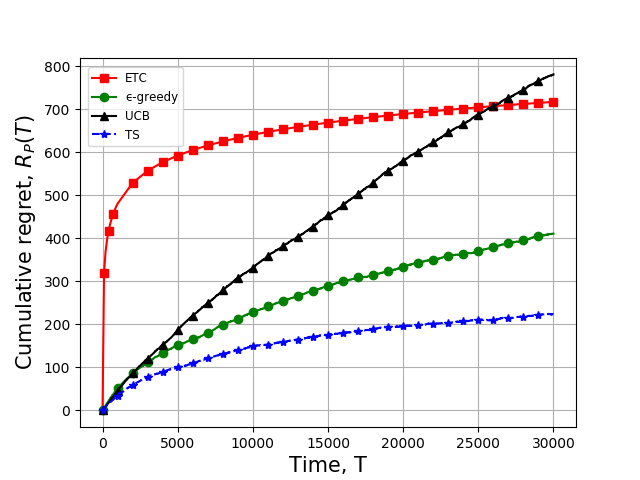}}
	\caption{(a) Instantaneous regret for all the four policies (ETC having $T_E = 13000$) for $K=5$, $p_k = [0.6, 0.65, 0.7, 0.75, 0.8]$, $q_k = [0.8, 0.75, 0.7, 0.65, 0.6]$, $d = 0.7$ (b)  Cumulative regret for all the four policies for $K=5$, $p_k = [0.6, 0.65, 0.7, 0.75, 0.8]$, $q_k = [0.8, 0.75, 0.7, 0.65, 0.6]$, $d = 0.7$}
	\label{2 plots: K=5}
\end{figure*}

\begin{figure*}[t]
	\centering     
	\subfigure[]{\label{i3}\includegraphics[width=80mm]{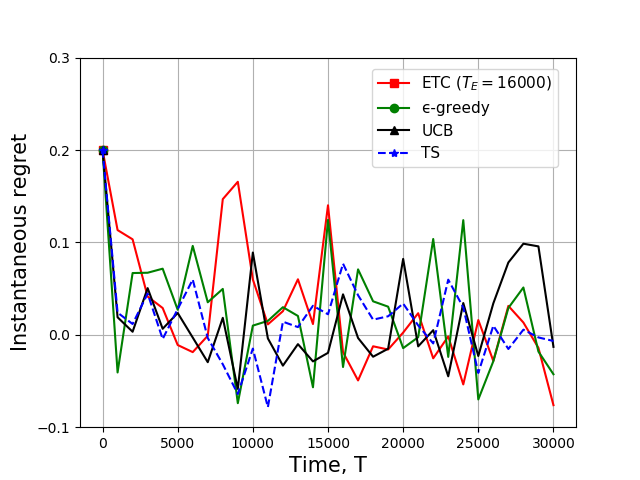}} 
	\subfigure[]{\label{c3}\includegraphics[width=80mm]{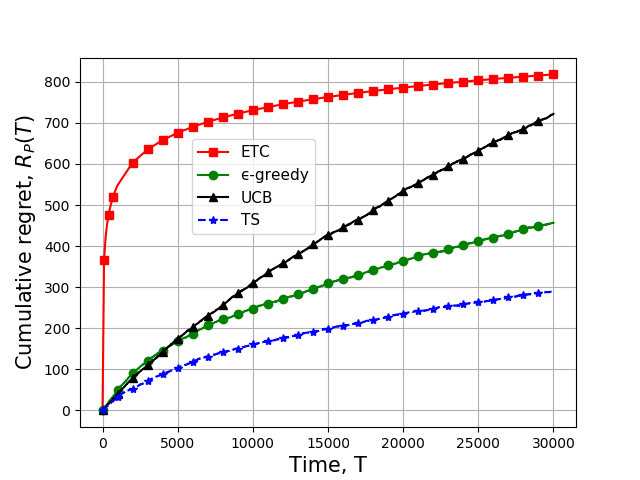}}
	\caption{(a) Instantaneous regret for all the four policies (ETC having $T_E = 16000$) for $K=7$, $p_k = [0.5, 0.55, 0.6, 0.65, 0.7, 0.75, 0.8]$, $q_k = [0.8, 0.75, 0.7, 0.65, 0.6, 0.55, 0.5]$, $d = 0.6$ (b)  Cumulative regret for all the four policies for $K=7$, $p_k = [0.5, 0.55, 0.6, 0.65, 0.7, 0.75, 0.8]$, $q_k = [0.8, 0.75, 0.7, 0.65, 0.6, 0.55, 0.5]$, $d = 0.6$}
	\label{2 plots: K=7}
\end{figure*}

\begin{figure}
	\centering     
    {\includegraphics[width=70mm]{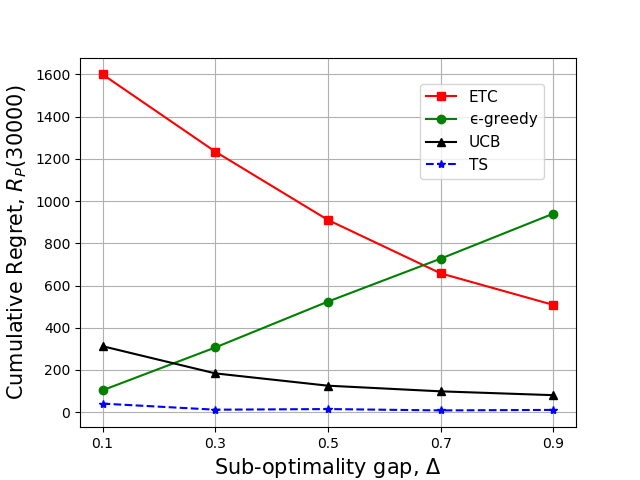}} 
	\caption{Cumulative regret as a function of $\Delta$ at time-slot $30000$ under the four policies}
 \label{varydelta}
\end{figure}

\begin{figure}
    \centering
    \vspace{-3mm}
    \subfigure[]{\label{varyp}\includegraphics[width=85mm]{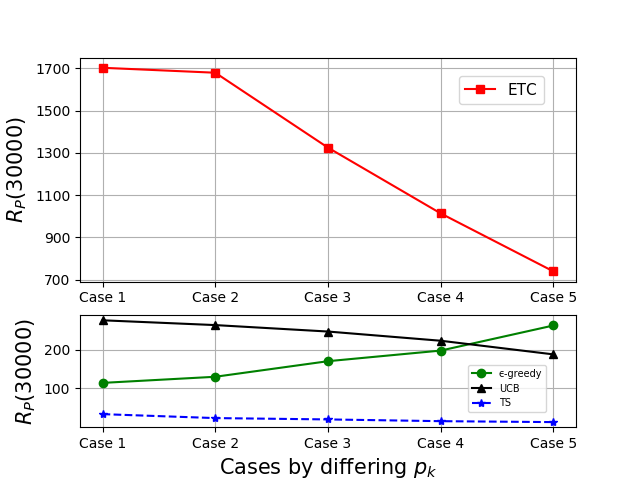}}
    \vspace{-3mm}
    \subfigure[]{\label{varyq}\includegraphics[width=85mm]{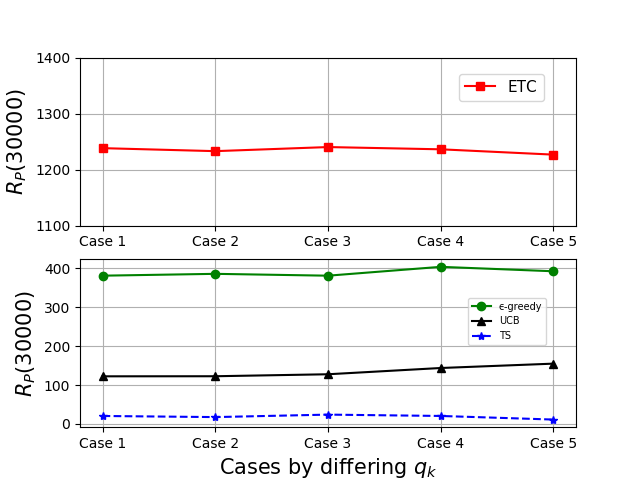}}
    \subfigure[]{\label{varyd}\includegraphics[width=85mm]{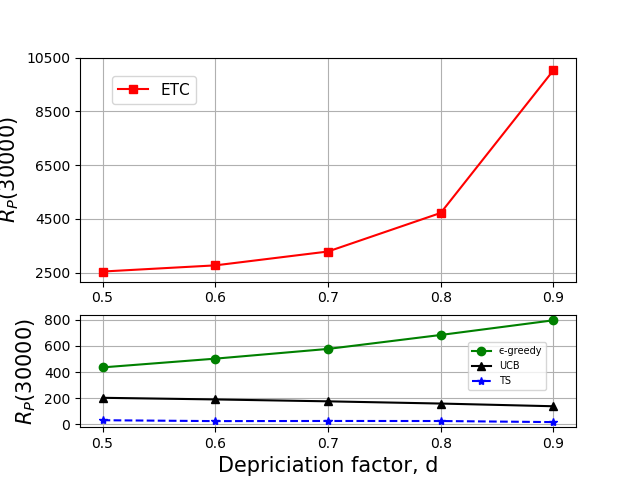}}
    \caption{(a)  Cumulative regret as a function of $p_k$ at time-slot $30000$ under the four policies (b)  Cumulative regret as a function of $q_k$ at time-slot $30000$ under the four policies (c) Cumulative regret as a function of $d$ at time slot $30000$ under the four policies}
    \label{fig:3 plots}
\end{figure}
In this section, we present our simulation results for all four policies discussed in section \ref{sec: scheduling policies}. All the simulations hereafter are conducted for $T = 3 \times 10^4$ time slots with each data point averaged across $200$ iterations.\\
\indent Recall that each policy has $K, p_k, q_k$ and $d$ as input parameters and ETC policy has an additional input parameter $T_E$. While $K$ is the number of sources, $p_k$ and $q_k$ are vectors denoting the probability of successful transmission and the probability of measuring the correct update for each of the sources. Further, $d$ is the depreciation factor and $T_E$ is the time of the exploration phase in the ETC policy. In Figure \ref{2 plots: K=4}, \ref{2 plots: K=5} and \ref{2 plots: K=7}, we plot the instantaneous regret, $\mathbb{E}[r^*(t) - r^\mathcal{P}(t)]$ $\forall$ $t = 1 ~\text{to}~ 30000$ and the corresponding cumulative regret, $\mathcal{R}_\mathcal{P}(T)~ \forall$~$T = 1$ to $30000$, for all the four policies for an instance where $K=4, 5$ and $7$ respectively and $p_k, q_k$ and $d$ are as given in the caption of the figures. For the ETC policy in all the $3$ cases, while the time evolution of the instantaneous regret is shown by taking the  $T_E$ values as $9000$, $13000$, and $16000$ respectively, the cumulative regret is plotted by varying the $T_E$ value for each time, $T$ appropriately (since $T_E$ depends on $T$) and plotting the corresponding regret value achieved at that time-slot, $T$.\\
\indent In all the $3$ cases, we have ensured by way of considering the values of $p_k$ and $q_k$ that the sources which have a higher probability of successful transmission have a lower probability of measuring the correct update. While this is not necessary under our model, it has indeed been chosen in this way to see the trade-off between the $p_k$ and $q_k$ values of the sources in deciding the optimal source and evolution of regret under this scenario.\\
\indent We observe that while the UCB policy achieves the highest regret at $T=30000$ for $K=4$, as the number of sources is increased, the cumulative regret of the ETC policy exceeds that of the UCB. In all the $3$ cases, we observe that the Thompson Sampling policy performs the best followed by the $\epsilon$-greedy policy. The performance analysis of the UCB and Thompson sampling policy remains an open problem. For ETC policy, while Theorem \ref{thm: ETC} gives $T_E$ values in the range of $10^5$ for higher values of $T$, it turns out that even for $T_E$ values as low as $9000$, ETC achieves a sub-linear regret.\\
\indent Figure \ref{varydelta} and \ref{fig:3 plots} shows the variation of cumulative regret values of all the $4$ policies at $T=30000$ with different parameters. Figure \ref{varydelta} indicates an inverse relationship of cumulative regret $\mathcal{R}_\mathcal{P}(30000)$ with $\Delta$, defined in \ref{thms}, for ETC, UCB and TS policies but scales linearly for the $\epsilon$-greedy policy. For ETC, UCB, TS policy, as $\Delta$ increases, the policy quickly gets to know the optimal source and starts scheduling the optimal source quite early in the run of the algorithm thus leading to its inverse relationship with $\Delta$. On the other hand, in the $\epsilon$-greedy policy, it schedules all the sub-optimal sources an almost equal number of time slots irrespective of the increasing $\Delta$ since the algorithm schedules the explore phase with probability $\epsilon_t$ which is independent of $\Delta$, thus, leading to its linear relationship with $\Delta$.\\
\indent For figure \ref{varyp}, $5$ cases have been considered by taking $2$ sources in the system. While the values of $q_k$ and $d$ have been kept constant for both the sources as $q_k = [0.8, 0.8]$ and $d=0.7$ in all the cases, the probability of successful transmission has been decreased for both the sources gradually as $\text{Case}~ 1: p_k = [0.9, 0.5]$, $\text{Case}~ 2: p_k = [0.8, 0.4]$, $\text{Case}~ 3: p_k = [0.7, 0.3]$, $\text{Case}~ 4: p_k = [0.6, 0.2]$, $\text{Case}~ 5: p_k = [0.5, 0.1]$. We observe that as cases proceed from $1$ to $5$, $\Delta$ of the system increases and hence a similar relationship is observed that cumulative regret has with $\Delta$ in figure \ref{varydelta}.\\
\indent Figure \ref{varyq} has been obtained by again considering $2$ sources in the system but here $p_k$ and $d$ have been kept constant for all the sources in all the cases as $p_k = [0.8, 0.8]$ and $d=0.7$ and $q_k$ values have been decreased for both the sources gradually over the $5$  cases in the similar way as the $p_k$ values decreased previously. We observe that the $\Delta$ value of the system is constant over all the cases and hence, the cumulative regret values also remain almost constant for all the cases for all the policies.\\
\indent Finally, we keep $p_k$ and $q_k$ values constant for both the sources as $p_k = [0.8, 0.5]$ and $q_k = [0.2, 0.8]$ and plot $\mathcal{R}_\mathcal{P}(30000)$ by varying the value of $d$ for obtaining the plot in figure \ref{varyd}. Again, we observe that $\Delta$ increases as $d$ increases and hence a similar relationship of cumulative regret with $d$ as it had with $\Delta$ for all the policies except ETC where we observe that as $d$ increases, $T_E$ value increases and hence the scheduling of the sub-optimal source increases and hence, $\mathcal{R}_\mathcal{P}(30000)$ increases.

\section{Conclusions}
\label{sec:conclusions}
We model a multi-source, single-channel setting where each source measures a time-varying quantity with varying levels of accuracy and one of them sends its update to the monitoring station. The probability that an attempted update is transmitted successfully depends on the source sending its update. We first characterize the Oracle policy. Then we propose $4$ standard bandit policies namely ETC, $\epsilon$-greedy, UCB, TS policy suitably adjusted to our system model where we steer the system towards using the optimal source. Then we give analytical guarantees of ETC and $\epsilon$-greedy policy. We demonstrate that the upper bound of cumulative regret of ETC and $\epsilon$- greedy policy scales as $O(K\log T)$ and $O(K^2\log ^4T)$. However, analytical guarantees for the other $2$ policies are still an open problem. Next via simulations, we compare the performance of all the $4$ policies and show that TS outperforms all the other policies. Finally, we provide a lower bound on the regret achievable by any policy.

\section{Acknowledgements}
\label{sec: acknowledgements}
I thank Prof. Sharayu Moharir, EE, IITB and Prof. Manjesh K. Hanawal, IEOR, IITB for their guidance throughout this work.

\bibliography{biblio}
\bibliographystyle{IEEEtran}

\section{appendix}
\label{sec: appendix}
In this section, we discuss the proof of all the theorems mentioned in the main paper. Please note that the numbering of equations in the appendix is in continuation with that in the main paper.
\subsection{Proof of Theorem \ref{thm1}}

\noindent \textbf{Theorem 1.} Given a problem instance $(p_k, q_k, d, K)$, the oracle policy uses the optimal source, $k^*$, given as
\begin{equation*}
    k^* = \arg \max_{1 \leq k \leq K}\Bigg(\frac{p_kq_k}{1-d(1-p_k)}\Bigg)
\end{equation*}
\begin{proof}
    Under the Oracle policy, we assume that the system starts at $t=-\infty$ and the scheduler schedules the optimal source in every time slot $t$.

    Let us assume that we schedule a source $k$, having the probability of successful transmission as $p_k$ and the probability of correct update as $q_k$, for transmission at every time slot $t$ starting from $t=-\infty$. Now, we calculate the expected value of the reward received by the monitoring station under this policy at any time $t$.

    Possible values of the reward received by the monitoring station at time $t = \{X, Xd, Xd^2, \hdots, \infty\}$ where $X \in \{0,1\}$ and $\mathbb{E}[X]=q_k$.
    \begin{flalign*}
        \mathbb{P}(\text{Reward} = X|X) &= p_k\\
        \mathbb{P}(\text{Reward} = Xd|X) &= p_k(1-p_k)\\
        \mathbb{P}(\text{Reward} = Xd^2|X) &= p_k(1-p_k)^2\\
        \vdots\\
        \mathbb{P}(\text{Reward} = Xd^n|X) &= p_k(1-p_k)^n\\
        \vdots
    \end{flalign*}
    \begin{flalign*}
        \mathbb{E}[\text{Reward}|X] &= Xp_k + Xdp_k(1-p_k) + Xd^2p_k(1-p_k)^2 + \hdots\\
        &\quad + Xd^np_k(1-p_k)^n + \hdots\\
        &= \sum_{n=0}^{\infty}Xd^np_k(1-p_k)^n
    \end{flalign*}
    \begin{flalign*}
        \mathbb{E}[\text{Reward}] &= \mathbb{E}_X[\mathbb{E}[\text{Reward}|X]]\\
        &= \mathbb{E}_X\Bigg[\sum_{n=0}^{\infty}Xd^np_k(1-p_k)^n\Bigg]\\
        &=\sum_{n=0}^{\infty}d^np_k(1-p_k)^n\mathbb{E}_X[X]\\
        &=\sum_{n=0}^{\infty}d^np_k(1-p_k)^nq_k\\
        &= p_kq_k\sum_{n=0}^{\infty}(d(1-p_k))^n\\
        &= \frac{p_kq_k}{1-d(1-p_k)}
    \end{flalign*}
    The optimal source would be the one in which the monitoring station receives the maximum reward upon scheduling it in all time slots starting from $t=-\infty$.

    Hence, the Optimal source is $k^*$, where,
    \begin{equation*}
        k^* = \arg \max\limits_{1 \leq k \leq K} \Bigg(\frac{p_kq_k}{1-d(1-p_k)}\Bigg)
    \end{equation*}
\end{proof}

\subsection{Proof of Theorem \ref{thm: ETC}}
\noindent \textbf{Theorem 2} (Performance of ETC policy). Let $p_{k^*}$ and $q_{k^*}$ be the probability of successful transmission and the probability of accurate measurement of the optimal source, $k^*$, as characterized in theorem \ref{thm1}. Then for $\alpha > 1$, $c = \max\Big\{\frac{-2}{\ln (d(1-p_{k^*}))},\frac{4K}{\Delta^2(1-d)^2}\Big\}$ and $T_E (\text{duration of the explore phase}) = c\ln T$ where, $\Delta = \mu_{k^*} - \max\limits_{k\neq k^*}\mu_k$,  $\mu_{k^*} = \frac{p_{k^*}q_{k^*}}{1-d(1-p_{k^*})}$ and $\mu_k = \frac{p_{k}q_{k}}{1-d(1-p_{k})}$,
\begin{equation*}
    \mathcal{R}_\mathcal{ETC}(T) \leq \frac{p_{k^*}q_{k^*}}{1-d(1-p_{k^*})}\Bigg[\alpha c\ln T \Bigg(1+\frac{K}{T^4}\Bigg)+\Bigg(\frac{K}{T}+\frac{1}{T^3}\Bigg)\Bigg]
\end{equation*}
\begin{proof}
    Let us assume a time slot $t > \alpha c\ln T$ where $\alpha > 1$. Note that we are considering a time slot in the commit phase since $T_E = c\ln T$. Further, let us assume source $k^*$ which is the optimal source as characterized by Theorem \ref{thm1} is scheduled in the time interval $(t-c\ln T,t]$. Note that the complete time interval considered $(t-c\ln T,t]$ is in the commit phase for the time slot $t$ assumed.
    \begin{align*}
    \mathbb{P}&(\text{source} ~k^*~ \text{is scheduled in the time interval} ~(t-c\ln T,t])\\
    &= \mathbb{P}(\text{source $k^*$ is scheduled in time slots $t-c\ln T+1$,}\\
    &\qquad ~~t-c\ln T+2, \hdots , t-1, t)\\
    &= \mathbb{P}(\text{source $k^*$ is scheduled in time slots $t$, $t-1$, $t-2$,}\\
    &\qquad ~ \hdots , t-c\ln T+1)\\
    &= \mathbb{P}(\text{source $k^*$ is scheduled in time slot $t$}~|~\text{source $k^*$ is}\\
    &\qquad ~~\text{scheduled in time slot $t-1$, $t-2$, $\hdots$, $t-c\ln T+1$})\\
    &\quad\times \mathbb{P}(\text{source $k^*$ is scheduled in time slot $t-1$}~|~\text{source $k^*$}\\
    &\qquad ~~~~\text{is scheduled in time slot $t-2$, $t-3$, $\hdots$ , $t-c\ln T+1$})\\
    &\quad\times \mathbb{P}(\text{source $k^*$ is scheduled in time slot $t-2$}~|~\text{source $k^*$}\\
    &\qquad ~~~~\text{is scheduled in time slot $t-3$, $t-4$, $\hdots$ , $t-c\ln T+1$})\\
    &~~~~~~~~~~~~~~~~~~~~~\vdots\\
    &~~~~~~~~~~~~~~~~~~~~~\vdots\\
    &\quad\times \mathbb{P}(\text{source $k^*$ is scheduled in time slot $t-c\ln T+2$}~|~\text{source}\\
    &\qquad~~~~\text{$k^*$ is scheduled in time slot $t-c\ln T+1$})\\
    &\quad\times\mathbb{P}(\text{source $k^*$ is scheduled in time slot $t-c\ln T+1$})
\end{align*}
Since the same source is scheduled after the explore phase,
\begin{align*}
    &\mathbb{P}(\text{source $k^*$ is scheduled in time slot $t$}~|~\text{source $k^*$ is}\\
    &\quad\text{scheduled in time slot $t-1$, $\hdots$, $t-c\ln T+1$}) = 1\\
    &\mathbb{P}(\text{source $k^*$ is scheduled in time slot $t-1$}~|~\text{source $k^*$ is}\\
    &\quad\text{scheduled in time slot $t-2$, $\hdots$ , $t-c\ln T+1$}) = 1\\
    &\mathbb{P}(\text{source $k^*$ is scheduled in time slot $t-2$}~|~\text{source $k^*$ is}\\
    &\quad\text{scheduled in time slot $t-3$, $\hdots$ , $t-c\ln T+1$}) = 1\\
    &~~~~~~~~~~~~~~~~~~~~~\vdots\\
    &~~~~~~~~~~~~~~~~~~~~~\vdots\\
    &\mathbb{P}(\text{source $k^*$ is scheduled in time slot $t-c\ln T+2$}~|\\
    &\quad\text{source $k^*$ is scheduled in time slot $t-c\ln T+1$}) = 1\\
\end{align*}
Let us now calculate $\mathbb{P}(\text{source $k^*$ is scheduled in time slot}$\\
\hspace*{3.5cm} $t-c\ln T+1$).\\

\noindent Consider,
\begin{align*}
    \mathbb{P}&(\text{source $k (\neq k^*)$ is scheduled after the explore phase})\\
    &= \mathbb{P}(\text{source $k (\neq k^*)$ is scheduled at the end of time slot}\\
    &\qquad ~~c\ln T)\\
    &=\mathbb{P}(\hat{\mu_k}(c\ln T) \geq \hat{\mu_{k^*}}(c\ln T))\\
    \end{align*}
    where,
    \begin{align*}
        \hat{\mu_k}(t) &= \text{Empirical estimate of $\mu_k$ for source $k$ till time $t$}\\
        &= \frac{\hat{PQ_k(t)}}{1-d(1-\hat{P_k}(t))}\\
        \hat{\mu_{k^*}}(t) &= \text{Empirical estimate of $\mu_{k^*}$ for source $k^*$ till time $t$}\\
        &= \frac{\hat{PQ_{k^*}(t)}}{1-d(1-\hat{P_{k^*}}(t))}
    \end{align*}
    where,
    \begin{align*}
        \hat{PQ_k(t)} &= \text{Empirical estimate of $p_kq_k$ for source $k$ till time}\\
        &~\quad\text{$t$ by using the local observations from the}\\
        &~\quad\text{monitoring station}\\
        &= \frac{\sum\limits_{s=1}^{t}\mathds{1}\{k(s)=k\}PQ_k(s)}{\sum\limits_{s=1}^{t}\mathds{1}\{k(s)=k\}}\\
        \hat{P_k(t)} &= \text{Empirical estimate of $p_k$ for source $k$ till time}\\
        &~\quad\text{$t$ by using the local observations from the}\\
        &~\quad\text{monitoring station}\\
        &= \frac{\sum\limits_{s=1}^{t}\mathds{1}\{k(s)=k\}P_k(s)}{\sum\limits_{s=1}^{t}\mathds{1}\{k(s)=k\}}\\
        \hat{PQ_{k^*}(t)} &= \text{Empirical estimate of $p_{k^*}q_{k^*}$ for source $k^*$ till}\\
        &~\quad\text{time $t$ by using the local observations from the}\\
        &~\quad\text{monitoring station}\\
        &= \frac{\sum\limits_{s=1}^{t}\mathds{1}\{k(s)=k^*\}PQ_{k^*}(s)}{\sum\limits_{s=1}^{t}\mathds{1}\{k(s)=k^*\}}\\
        \hat{P_{k^*}(t)} &= \text{Empirical estimate of $p_{k^*}$ for source $k^*$ till time}\\
        &~\quad\text{$t$ by using the local observations from the}\\
        &~\quad\text{monitoring station}\\
        &= \frac{\sum\limits_{s=1}^{t}\mathds{1}\{k(s)=k^*\}P_{k^*}(s)}{\sum\limits_{s=1}^{t}\mathds{1}\{k(s)=k^*\}}
    \end{align*}
    where,
    \begin{align*}
        PQ_k(s) &= \text{sample of $PQ_k$ for source $k$ at time slot $s$}\\
        P_k(s) &= \text{sample of $P_k$ for source $k$ at time slot $s$}\\
        PQ_{k^*}(s) &= \text{sample of $PQ_{k^*}$ for source $k^*$ at time slot $s$}\\
        P_{k^*}(s) &= \text{sample of $P_{k^*}$ for source $k^*$ at time slot $s$}
    \end{align*}
    Then,
    \begin{align*}
    \mathbb{P}&(\text{source $k (\neq k^*)$ is scheduled after the explore phase})\\
    &=\mathbb{P}(\hat{\mu_k}(c\ln T) \geq \hat{\mu_{k^*}}(c\ln T))\\
    &= \mathbb{P}(\hat{\mu_k}(c\ln T)-\hat{\mu_{k^*}}(c\ln T)\geq 0)\\
    &= \mathbb{P}((\hat{\mu_k}(c\ln T)- \mu_k)-(\hat{\mu_{k^*}}(c\ln T)-\mu_{k^*})\geq \mu_{k^*}-\mu_k)\\
    &= \mathbb{P}((\hat{\mu_k}(c\ln T)- \mu_k)-(\hat{\mu_{k^*}}(c\ln T)-\mu_{k^*})\geq \Delta_k)\\
    &~~~~~~~~~~~~~~~~~~~~~~~~~~~~~~\qquad\qquad \text{where ,} ~\Delta_k = \mu_{k^*} - \mu_k\\
    &\leq \mathbb{P}((\hat{\mu_k}(c\ln T)- \mu_k)-(\hat{\mu_{k^*}}(c\ln T)-\mu_{k^*})\geq \Delta)\\
    &~~~~~~~~~~~~~~~~~~~~~~~\text{where ,}~\Delta = \min\limits_{k\neq k^*}\Delta_k = \mu_{k^*} - \max\limits_{k\neq k^*}\mu_k\\
    &\leq \exp\Bigg(\frac{-2\frac{c\ln T}{K}\Delta^2}{(\frac{2}{1-d})^2}\Bigg)~~~~~~~~~~~~~~\text{(Hoeffding's inequality)}\\
    &= \exp\Bigg(\frac{-c\ln T\Delta^2(1-d)^2}{2K}\Bigg)\\
\end{align*}
Hence,
\begin{align*}
    \mathbb{P}&(\text{source $k^*$ is scheduled after the explore phase})\\
    &\geq 1- \sum_{\substack{k=1 \\~ i\neq k^*}}^{K}\exp\Bigg(\frac{-c\ln T\Delta^2(1-d)^2}{2K}\Bigg)\\
    &= 1- (K-1)\exp\Bigg(\frac{-c\ln T\Delta^2(1-d)^2}{2K}\Bigg)
\end{align*}
Hence,
\begin{align*}
    \mathbb{P}&(\text{source $k^*$ is scheduled in time slot $t-c\ln T+1$})\\
    & \geq 1- (K-1)\exp\Bigg(\frac{-c\ln T\Delta^2(1-d)^2}{2K}\Bigg)
\end{align*}
Hence,
\begin{align}
    \mathbb{P}&(\text{source $k^*$ is scheduled in the time interval $(t-c\ln T,t]$})\nonumber\\
    &\geq 1- (K-1)\exp\Bigg(\frac{-c\ln T\Delta^2(1-d)^2}{2K}\Bigg)
\end{align}
Let us now calculate the expected reward received by the monitoring station under the ETC policy conditioned on the event that it schedules the optimal source $k^{*}$ in the time interval $(t-c\ln T,t]$.
\begin{align}
    \mathbb{E}&[r^\mathcal{ETC}(t)~|~\text{source  $k^*$ is scheduled in time interval}\nonumber\\
    &~\text{$(t-c\ln T,t]$}]\nonumber\\
    &= q_{k^*}p_{k^*}+q_{k^*}dp_{k^*}(1-p_{k^*}) +q_{k^*}d^2p_{k^*}(1-p_{k^*})^2\nonumber\\
    &\quad+ \hdots \hdots + q_{k^*}d^{c\ln T-1}p_{k^*}(1-p_{k^*})^{c\ln T-1} + \hdots\hdots\nonumber\\
    &\geq q_{k^*}p_{k^*}+q_{k^*}dp_{k^*}(1-p_{k^*})+q_{k^*}d^2p_{k^*}(1-p_{k^*})^2\nonumber\\
    &\quad+ \hdots \hdots + q_{k^*}d^{c\ln T-1}p_{k^*}(1-p_{k^*})^{c\ln T-1}\nonumber\\
    &= p_{k^*}q_{k^*}(1+d(1-p_{k^*})+d^2(1-p_{k^*})^2+ \hdots \hdots\nonumber \\
    &\quad + d^{c\ln T-1}(1-p_{k^*})^{c\ln T-1})\nonumber\\
    &= p_{k^*}q_{k^*}\Bigg[\frac{1-(d(1-p_{k^*}))^{c\ln T}}{1-d(1-p_{k^*})}\Bigg]
\end{align}
Let us now calculate the lower bound on the expected reward under ETC policy at any time slot $t$.
\begin{align*}
    \mathbb{E}[r^\mathcal{ETC}(t)] &= \mathbb{E}[r^\mathcal{ETC}(t)~|~\text{source  $k^*$ is scheduled in time}\\
    &~~~~\quad\text{interval $(t-c\ln T,t]$}]~\mathbb{P}(\text{source $k^*$ is}\\
    &~~~~~\quad\text{scheduled in time interval $(t-c\ln T,t]$})\\
    &\quad+ \mathbb{E}[r^\mathcal{ETC}(t)~|~\text{source $k(\neq k^*)$ is scheduled in}\\
    &~~~~~~~\quad\text{time interval $(t-c\ln T,t]$}]~\mathbb{P}(\text{source $k(\neq k^*)$}\\
    &~~~~~~~\quad\text{is scheduled in time interval $(t-c\ln T,t]$})\\
    &\geq \mathbb{E}[r^\mathcal{ETC}(t)~|~\text{source  $k^*$ is scheduled in time}\\
    &~~~~\quad\text{interval $(t-c\ln T,t]$}]~\mathbb{P}(\text{source $k^*$ is}\\
    &~~~~~\quad\text{scheduled in time interval $(t-c\ln T,t]$})\\
\end{align*}
Substituting from Equation $(2)$ and $(3)$ above, we get
\begin{align*}
    \mathbb{E}[r^\mathcal{ETC}(t)] &\geq p_{k^*}q_{k^*}\Bigg[\frac{1-(d(1-p_{k^*}))^{c\ln T}}{1-d(1-p_{k^*})}\Bigg]\\
    &\quad\times\Bigg(1- (K-1)\exp\Bigg(\frac{-c\ln T\Delta^2(1-d)^2}{2K}\Bigg)\Bigg)
\end{align*}
Let us calculate the cumulative regret of the system from time slot $1$ to $T$. Since we want to get an upper bound of the cumulative regret and check how it scales with time, let us assume we do not accumulate any reward from time slot $t=1$ to $t=\alpha c\ln T$, where, $\alpha > 1$.
\begin{align*}
    \mathcal{R}_\mathcal{ETC}(T) &=\sum_{t=1}^{T}\mathbb{E}[r^*(t) -r^\mathcal{ETC}(t)]\\
    &\leq \sum_{t=1}^{\alpha c\ln T}\Bigg(\frac{p_{k^*}q_{k^*}}{1-d(1-p_{k^*})} - 0\Bigg)\\
    &\quad + \sum_{t=\alpha c\ln T+1}^{T}\Bigg[\frac{p_{k^*}q_{k^*}}{1-d(1-p_{k^*})}\\
    &\quad -p_{k^*}q_{k^*}\Bigg[\frac{1-(d(1-p_{k^*}))^{c\ln T}}{1-d(1-p_{k^*})}\Bigg]\\
    &\quad\times\Bigg(1- (K-1)\exp\Bigg(\frac{-c\ln T\Delta^2(1-d)^2}{2K}\Bigg)\Bigg)\Bigg]\\
    &= \frac{p_{k^*}q_{k^*}(\alpha c\ln T)}{1-d(1-p_{k^*})} + \sum_{t=\alpha c\ln T+1}^{T}\Bigg[\frac{p_{k^*}q_{k^*}}{1-d(1-p_{k^*})}\\
    &\quad -p_{k^*}q_{k^*}\Bigg[\frac{1-(d(1-p_{k^*}))^{c\ln T}}{1-d(1-p_{k^*})}\Bigg]\\
    &\quad\times\Bigg(1- (K-1)\exp\Bigg(\frac{-c\ln T\Delta^2(1-d)^2}{2K}\Bigg)\Bigg)\Bigg]\\
    &= \frac{p_{k^*}q_{k^*}(\alpha c\ln T)}{1-d(1-p_{k^*})} + \frac{p_{k^*}q_{k^*}}{1-d(1-p_{k^*})}\sum_{t=\alpha c\ln T+1}^{T}\Bigg[1\\
    &\quad - \Big(1-(d(1-p_{k^*}))^{c\ln T}\Big)\\
    &\quad\times\Bigg(1- (K-1)\exp\Bigg(\frac{-c\ln T\Delta^2(1-d)^2}{2K}\Bigg)\Bigg)\Bigg]\\
\end{align*}
Now, since
\begin{flalign*}
    c&=\max\Bigg\{\frac{-2}{\ln (d(1-p_{k^*}))},\frac{4K}{\Delta^2(1-d)^2}\Bigg\}\\
\implies c &\geq \frac{-2}{\ln (d(1-p_{k^*}))}
\end{flalign*}
Consider
\begin{align*}
    (d(1-p_{k^*}))^{c \ln T} &= \exp{(\ln (d(1-p_{k^*}))^{c\ln T})}\\
    &= \exp{(c\ln T\ln (d(1-p_{k^*})))}\\
    &\leq \exp{\Bigg(\frac{-2}{\ln (d(1-p_{i^*}))}\ln T \ln (d(1-p_{k^*}))\Bigg)}\\
    &= \exp{(-2 \ln T)}\\
    &= \exp{(\ln T^{-2})}\\
    &= \frac{1}{T^2}
\end{align*}
Again, since
\begin{flalign*}
    c&=\max\Bigg\{\frac{-2}{\ln (d(1-p_{i^*}))},\frac{4K}{\Delta^2(1-d)^2}\Bigg\}\\
\implies c &\geq \frac{4K}{\Delta^2(1-d)^2}\\
\implies& \frac{c\Delta^2(1-d)^2}{2K} \geq 2
\end{flalign*}
Consider
\begin{align*}
    \exp\Bigg(\frac{-c\ln T\Delta^2(1-d)^2}{2K}\Bigg) &\leq \exp{(-2 \ln T)}\\
    &= \exp{(\ln T^{-2})}\\
    &= \frac{1}{T^2}
\end{align*}
Therefore,
\begin{align*}
    \mathcal{R}_\mathcal{ETC}(T) &\leq \frac{p_{k^*}q_{k^*}(\alpha c\ln T)}{1-d(1-p_{k^*})} + \frac{p_{k^*}q_{k^*}}{1-d(1-p_{k^*})}\sum_{t=\alpha c\ln T+1}^{T}\Bigg[1\\
    &\quad - \Big(1-\frac{1}{T^2}\Big)\Bigg(1- (K-1)\times\frac{1}{T^2}\Bigg)\Bigg]\\
    &= \frac{p_{k^*}q_{k^*}(\alpha c\ln T)}{1-d(1-p_{k^*})} + \frac{p_{k^*}q_{k^*}}{1-d(1-p_{k^*})}\sum_{t=\alpha c\ln T+1}^{T}\Bigg[1\\
    &\quad - \Bigg(1- \frac{K-1}{T^2} - \frac{1}{T^2} + \frac{K-1}{T^4}\Bigg)\Bigg]\\
    &= \frac{p_{k^*}q_{k^*}(\alpha c\ln T)}{1-d(1-p_{k^*})} + \frac{p_{k^*}q_{k^*}}{1-d(1-p_{k^*})}\sum_{t=\alpha c\ln T+1}^{T}\\
    &\quad\Bigg(\frac{K-1}{T^2} + \frac{1}{T^2} - \frac{K-1}{T^4}\Bigg)\\
    &= \frac{p_{k^*}q_{k^*}}{1-d(1-p_{k^*})}\Bigg[\alpha c\ln T + \sum_{t=\alpha c\ln T+1}^{T}\Bigg(\frac{K-1}{T^2}\\
    &\quad + \frac{1}{T^2} - \frac{K-1}{T^4}\Bigg)\Bigg]\\
    &= \frac{p_{k^*}q_{k^*}}{1-d(1-p_{k^*})}\Bigg[\alpha c\ln T + \Bigg(\frac{K}{T^2} - \frac{K-1}{T^4}\Bigg)\\
    &\quad\times(T-\alpha c\ln T)\Bigg]\\
    &= \frac{p_{k^*}q_{k^*}}{1-d(1-p_{k^*})}\Bigg[\alpha c\ln T \Bigg(1+\frac{K}{T^4}\Bigg)+\Bigg(\frac{K}{T}\\
    &\quad +\frac{1}{T^3}\Bigg) -\frac{\alpha c\ln T}{T^2}\Bigg(K+\frac{1}{T^2}\Bigg)-\frac{K}{T^3}\Bigg]\\
    &\leq \frac{p_{k^*}q_{k^*}}{1-d(1-p_{k^*})}\Bigg[\alpha c\ln T \Bigg(1+\frac{K}{T^4}\Bigg)+\Bigg(\frac{K}{T}+\frac{1}{T^3}\Bigg)\Bigg]\\
\end{align*}
\end{proof}

\subsection{Proof of Theorem \ref{thm: epsilon greedy}}
\textbf{Theorem 3} (Performance of $\epsilon$- greedy). Let $p_{k^*}$ and $q_{k^*}$ be the probability of successful transmission and the probability of accurate measurement of the optimal source, $k^*$, as characterized in theorem \ref{thm1}. Then for $\alpha > 1$, $c = \max\Big\{\frac{-2}{\ln (d(1-p_{k^*}))},\frac{4K}{\Delta^2(1-d)^2}\Big\}$ and $\epsilon_t = \min\Big\{1,3K\frac{\ln^2 t}{t}\Big\}$ where, $\Delta = \mu_{k^*} - \max\limits_{k\neq k^*}\mu_k$,  $\mu_{k^*} = \frac{p_{k^*}q_{k^*}}{1-d(1-p_{k^*})}$ and $\mu_k = \frac{p_{k}q_{k}}{1-d(1-p_{k})}$,
\begin{align*}
    \mathcal{R}_\epsilon(T)&\leq \frac{p_{k^*}q_{k^*}}{1-d(1-p_{k^*})}\Bigg[\alpha c\ln T + (K-1)\\
    &\quad\times\frac{\frac{c}{K}}{\Big(1-\frac{c}{K}\Big)}\Bigg(\frac{1}{(\alpha-1)c\ln T+1}\Bigg)^{\frac{1-c/K}{c/K}}\\
    &\quad + Kc\ln T\ln^3(T-c\ln T+1)\\
    &\quad + \frac{1}{T} +\frac{c^2\ln T (K-1)}{T^2}\\
    &\quad\times\exp\Bigg({\frac{-\ln^3(\alpha c\ln T-c\ln T+1)}{\frac{c}{K}}}\Bigg)\Bigg]
\end{align*}
\begin{proof}
    We assume that the monitoring station schedules a source uniformly at random in time slot $t$ (Exploration) with probability $\epsilon_t = \min\Big\{1,3K\frac{\ln^2 t}{t}\Big\}$ and schedules the empirically best source till time slot $t$ (Exploitation) with probability $(1-\epsilon_t)$.

    Let us assume a time slot $t > \alpha c\ln T$ where $\alpha > 1$. Further, let us assume source $k^*$ which is the optimal source as characterized by Theorem \ref{thm1} is scheduled by exploitation in the time interval $(t-c\ln T,t]$. It implies that in the complete time interval $(t-c\ln T,t]$, source $k^*$ is assumed to be the empirically best source.
    \begin{align*}
        \mathbb{P}&(\text{source $k^*$ is exploited in the time interval $(t-c \ln T,t]$})\\
     &= \mathbb{P}(\text{source $k^*$ is exploited in time slots $t-c\ln T+1$,}\\
     &~\qquad\text{$t-c\ln T+2$, $\hdots$ , $t-1$, $t$)}\\
    &= \mathbb{P}(\text{source $k^*$ is exploited in time slots $t$, $t-1$, $t-2$, $\hdots$,}\\
    &\qquad~~\text{$t-c\ln T+1$)}\\
    &= \mathbb{P}(\text{source $k^*$ is exploited in time slot $t$}~|~\text{source $k^*$ is}\\
    &\qquad~~\text{exploited in time slot $t-1$, $t-2$, $\hdots$, $t-c\ln T+1$})\\
    &~~~~\times \mathbb{P}(\text{source $k^*$ is exploited in time slot $t-1$}~|~\text{source $k^*$ is}\\
    &\qquad~~~~~\text{exploited in time slot $t-2$, $\hdots$ , $t-c\ln T+1$})\\
    &~~~~\times \mathbb{P}(\text{source $k^*$ is exploited in time slot $t-2$}~|~\text{source $k^*$ is}\\
    &\qquad~~~~~\text{exploited in  time slot $t-3$, $\hdots$ , $t-c\ln T+1$})\\
    &~~~~~~~~~~~~~~~~~~~~~\vdots\\
    &~~~~~~~~~~~~~~~~~~~~~\vdots\\
    &~~~~\times \mathbb{P}(\text{source $k^*$ is exploited in time slot $t-c\ln T+2$}~|~\\
    &\qquad~~~~~\text{source $k^*$ is exploited in time slot $t-c\ln T+1$})\\
    &~~~~\times\mathbb{P}(\text{source $k^*$ is exploited in time slot $t-c\ln T+1$})
    \end{align*}
    Let us now calculate each of the terms above.\\
    
    Consider,
    \begin{align*}
        \mathbb{P}&(\text{source $k^*$ is exploited in time slot $t-c\ln T+2$}~|~\text{source $k^*$}\\
        &~~\text{is exploited in time slot $t-1$, $t-2$, $\hdots$, $t-c\ln T+1$})\\
        &= \mathbb{P}(\text{source $k^*$ is scheduled in $t-c\ln T +2$}~|~\text{source $k^*$ is}\\
        &~\qquad\text{exploited in $t-c\ln T +1$, exploit in time slot}\\
        &~\qquad\text{$t-c\ln T +2$}) \times\mathbb{P}(\text{exploit in time slot $t-c\ln T+2$})\\
        &= 1\times\mathbb{P}(\text{exploit in time slot $t-c\ln T+2$})\\
        &= 1-\epsilon_{t-c\ln T+2}
    \end{align*}
    The conditional probability in the second step is $1$ because it is assumed in the implementation of this policy that in the exploit time slot, the empirical estimates of any of the sources will not be updated in spite of the fact that the monitoring station receives the local observation of the source that it selects in that time slot.
    
    So, if time slot $t$ is an exploit time slot in which source $k^*$ is scheduled for sending its updates, it suggests that
    \begin{align*}
        \hat{\mu_{k^*}}(t) > \hat{\mu_{k}}(t) ~\forall~ k\neq k^*
    \end{align*}
    Here $\hat{\mu_{k^*}}(t)$ and $\hat{\mu_{k}}(t)$ are as defined in the proof of Theorem \ref{thm: ETC}.
    Also, empirical estimates of any of the sources are not updated in time slot $t$ (since $t$ is an exploit time slot), and hence, source $k^*$ is again bound to get scheduled in time slot $t+1$ in case it is also an exploit time slot.\\
    Similarly,
    \begin{align*}
    \mathbb{P}&(\text{source $k^*$ is exploited in time slot $t-c\ln T+3$}~|~\text{source $k^*$}\\
    &~~\text{is exploited in time slot $t-c\ln T +2, t-c\ln T+1$})\\
    &= \mathbb{P}(\text{source $k^*$ is scheduled in time slot $t-c\ln T +3$}~|\\
    &~\qquad\text{source $k^*$ is exploited in $t-c\ln T +2,~ t-c\ln T+1$,}\\
    &~\qquad\text{exploit in time slot $t-c\ln T +3$})\\
    &\quad\times\mathbb{P}(\text{exploit in time slot $t-c\ln T+3$})\\
    &= 1 \times\mathbb{P}(\text{exploit in time slot $t-c\ln T+3$})\\
    &= 1-\epsilon_{t-c\ln T+3}
    \end{align*}
    The conditional probability in the second step is $1$ because of the same reason as mentioned above.
    \begin{flalign*}
    &~~~~\vdots\\
    &~~~~\vdots\\
    \end{flalign*}
    \begin{flalign*}
        \mathbb{P}&(\text{source $k^*$ is exploited in time slot $t$}~|~\text{source $k^*$ is exploited}\\
        &~~\text{in time slot $t-1, t-2, \hdots, t-c\ln T+1$})\\
        &= \mathbb{P}(\text{source $k^*$ is scheduled in time slot $t$}~|~\text{source $k^*$ is}\\
        &~~\qquad\text{exploited in $t-1, t-2, \hdots, t-c\ln T+1$, exploit}\\
        &~~\qquad\text{in time slot $t$}) \times\mathbb{P}(\text{exploit in time slot $t$})\\
        &= 1 \times\mathbb{P}(\text{exploit in time slot $t$})\\
        &= 1-\epsilon_{t}
    \end{flalign*}
    Hence,
    \begin{flalign*}
        \mathbb{P}&(\text{source $k^*$ is exploited in the time interval $(t-c\ln T,t]$})\\
        &= (1-\epsilon_{t-c\ln T+2})(1-\epsilon_{t-c\ln T+3})\hdots(1-\epsilon_{t-1})(1-\epsilon_{t})\\
    &\quad\times\mathbb{P}(\text{source $k^*$ is exploited in time slot $t-c\ln T+1$})
    \end{flalign*}
Let us now calculate $\mathbb{P}(\text{source $k^*$ is exploited in time slot}$\\
\hspace*{3.4cm} $t-c\ln T+1$).\\
Consider,
\begin{flalign*}
    \mathbb{P}&(\text{source $k^*$ is exploited in time slot $t$})\\
    &= \mathbb{P}(\text{source $k^*$ is scheduled in time slot $t$}~|\\
    &~~\qquad\text{exploit in time slot $t$})\times\mathbb{P}(\text{exploit in time slot $t$})\\
    &= \mathbb{P}(\text{source $k^*$ is scheduled in time slot $t$}~|\\
    &~~\qquad\text{exploit in time slot $t$})\times (1-\epsilon_t)
\end{flalign*}
Consider,
\begin{flalign*}
    \mathbb{P}&(\text{source $k (\neq k^*)$ is scheduled in time slot $t$}~|~\text{exploit in time slot $t$})\\
    &= \mathbb{P}(\hat{\mu_k}(t) \geq \hat{\mu_{k^*}}(t))\\
    &\text{(where, $\hat{\mu_k}(t)$ and $\hat{\mu_{k^*}}(t)$ are as defined in the proof of Theorem \ref{thm: ETC})}\\
    &= \mathbb{P}(\hat{\mu_k}(t) - \hat{\mu_{k^*}}(t) \geq 0)\\
    &= \mathbb{P}((\hat{\mu_k}(t)- \mu_k)-(\hat{\mu_{k^*}}(t)-\mu_{k^*})\geq \mu_{k^*}-\mu_k)\\
    &= \mathbb{P}((\hat{\mu_k}(t)- \mu_k)-(\hat{\mu_{k^*}}(t)-\mu_{k^*})\geq \Delta_k)\\
    &~~~~~~~~~~~~~~~~~~~~~~~~~~~~~~\qquad\qquad \text{where ,} ~\Delta_k = \mu_{k^*} - \mu_k\\
    &\leq \mathbb{P}((\hat{\mu_k}(t)- \mu_k)-(\hat{\mu_{k^*}}(t)-\mu_{k^*})\geq \Delta)\\
    &~~~~~~~~~~~~~~~~~~~~~~~\text{where ,}~\Delta = \min\limits_{k\neq k^*}\Delta_k = \mu_{k^*} - \max\limits_{k\neq k^*}\mu_k\\
    &\leq \exp\Bigg(\frac{-2\Big(\frac{1}{2}\ln^3(t)\Big)\Delta^2}{(\frac{2}{1-d})^2}\Bigg)~~~~~~~~\text{(Hoeffding's inequality)}\\
    &= \exp\Bigg(\frac{-\ln^3 (t)\Delta^2(1-d)^2}{4}\Bigg)\\
\end{flalign*} 
Last but one inequality is because of the fact that $u_k(t) \geq \frac{1}{2}\ln^3(t)$, where, $u_k(t) = $ number of times source $k$ has been scheduled till time slot $t$ \cite{krishnasamy2016regret}.\\

\noindent Hence,
\begin{flalign*}
    \mathbb{P}&(\text{source $k^*$ is scheduled in time slot $t$}~|~\text{exploit in time slot $t$})\\
    &\geq 1-\sum_{\substack{k=1 \\~ k\neq k^*}}^{K}\exp\Bigg(\frac{-\ln^3 (t)\Delta^2(1-d)^2}{4}\Bigg)\\
    &= 1-(K-1)\exp\Bigg(\frac{-\ln^3 (t)\Delta^2(1-d)^2}{4}\Bigg)
\end{flalign*}
Hence,
\begin{flalign*}
    \mathbb{P}&(\text{source $k^*$ is exploited in time slot $t$})\\
    &\geq \Bigg[1-(K-1)\exp\Bigg(\frac{-\ln^3 (t)\Delta^2(1-d)^2}{4}\Bigg)\Bigg]\times (1-\epsilon_t)
\end{flalign*}
Hence,
\begin{flalign*}
    \mathbb{P}&(\text{source $k^*$ is exploited in time slot $t-c\ln T +1$})\\
    &\geq \Bigg[1-(K-1)\exp\Bigg(\frac{-\ln^3 (t-c\ln T+1)\Delta^2(1-d)^2}{4}\Bigg)\Bigg]\\
    &\quad\times (1-\epsilon_{t-c\ln T+1})
\end{flalign*}
Hence,
\begin{flalign}
    \mathbb{P}&(\text{source $k^*$ is exploited in the time interval $(t-c\ln T,t]$})\nonumber\\
        &\geq (1-\epsilon_{t-c\ln T+2})(1-\epsilon_{t-c\ln T+3})\hdots(1-\epsilon_{t-1})(1-\epsilon_{t})\nonumber\\
        &\quad\times\Bigg[1-(K-1)\exp\Bigg(\frac{-\ln^3 (t-c\ln T+1)\Delta^2(1-d)^2}{4}\Bigg)\Bigg]\nonumber\\
        &\quad\times (1-\epsilon_{t-c\ln T+1})\nonumber\\
        &=\prod_{s=t-c\ln T+1}^{t}(1-\epsilon_s)\nonumber\\
    &\quad\times\Bigg[1-(K-1)\exp\Bigg(\frac{-\ln^3 (t-c\ln T+1)\Delta^2(1-d)^2}{4}\Bigg)\Bigg]
\end{flalign}
The expected reward received by the monitoring station under the $\epsilon$- greedy policy conditioned on the event that it exploits the optimal source $k^*$ in the time interval $(t-c\ln T,t]$ is the same as the expected reward under the ETC policy conditioned on the same event.\\
Hence,
\begin{flalign*}
    \mathbb{E}&[r^\mathcal{\epsilon}(t)~|~\text{source  $k^*$ is exploited in the time interval}\\
    &~\text{$(t-c\ln T,t]$}]\\
    &= p_{k^*}q_{k^*}\Bigg[\frac{1-(d(1-p_{k^*}))^{c\ln T}}{1-d(1-p_{k^*})}\Bigg]~~~~~~~~~~~~~~~~~~~~~~~~~(3)
\end{flalign*}
Let us now calculate the lower bound on the expected reward under $\epsilon$- greedy policy at any time slot $t$.
\begin{flalign*}
    \mathbb{E}[r^\mathcal{\epsilon}(t)] &= \mathbb{E}[r^\mathcal{\epsilon}(t)~|~\text{source  $k^*$ is exploited in time}\\
    &~~~~\quad\text{interval $(t-c\ln T,t]$}]~\mathbb{P}(\text{source $k^*$ is}\\
    &~~~~~\quad\text{exploited in time interval $(t-c\ln T,t]$})\\
    &\quad+ \mathbb{E}[r^\mathcal{\epsilon}(t)~|~\text{source $k(\neq k^*)$ is exploited in}\\
    &~~~~~~~\quad\text{time interval $(t-c\ln T,t]$}]~\mathbb{P}(\text{source $k(\neq k^*)$}\\
    &~~~~~~~\quad\text{is exploited in time interval $(t-c\ln T,t]$})\\
    &\geq \mathbb{E}[r^\mathcal{\epsilon}(t)~|~\text{source  $k^*$ is exploited in time}\\
    &~~~~\quad\text{interval $(t-c\ln T,t]$}]~\mathbb{P}(\text{source $k^*$ is}\\
    &~~~~~\quad\text{exploited in time interval $(t-c\ln T,t]$})\\
\end{flalign*}
Substituting from Equation $(3)$ and $(4)$, we get
\begin{flalign*}
    \mathbb{E}[r^\mathcal{\epsilon}(t)] &\geq p_{k^*}q_{k^*}\Bigg[\frac{1-(d(1-p_{k^*}))^{c\ln T}}{1-d(1-p_{k^*})}\Bigg]\\
    &\quad\times\Bigg(\prod_{s=t-c\ln T+1}^{t}(1-\epsilon_s)\Bigg) \times\Bigg[1-(K-1)\\
    &\quad\times\exp\Bigg(\frac{-\ln^3 (t-c\ln T+1)\Delta^2(1-d)^2}{4}\Bigg)\Bigg]\\
    &\geq p_{k^*}q_{k^*}\Bigg[\frac{1-(d(1-p_{k^*}))^{c\ln T}}{1-d(1-p_{k^*})}\Bigg]\\
    &\quad\times\Bigg(\prod_{s=t-c\ln T+1}^{t}\Big(1-3K\frac{\ln^2 s}{s}\Big)\Bigg) \times\Bigg[1-(K-1)\\
    &\quad\times\exp\Bigg(\frac{-\ln^3 (t-c\ln T+1)\Delta^2(1-d)^2}{4}\Bigg)\Bigg]\\
    &\geq p_{k^*}q_{k^*}\Bigg[\frac{1-(d(1-p_{k^*}))^{c\ln T}}{1-d(1-p_{k^*})}\Bigg]\\
    &\quad\times\Bigg(1-3K\frac{\ln^2 (t-c\ln T+1)}{t-c\ln T+1}\Bigg)^{c\ln T}\times\Bigg[1-(K-1)\\
    &\quad\times\exp\Bigg(\frac{-\ln^3 (t-c\ln T+1)\Delta^2(1-d)^2}{4}\Bigg)\Bigg]\\
    &\geq p_{k^*}q_{k^*}\Bigg[\frac{1-(d(1-p_{k^*}))^{c\ln T}}{1-d(1-p_{k^*})}\Bigg]\\
    &\quad\times\Bigg(1-3K\frac{c\ln T\ln^2 (t-c\ln T+1)}{t-c\ln T+1}\Bigg)\times\Bigg[1\\
    &\quad -(K-1)\times\exp\Bigg(\frac{-\ln^3 (t-c\ln T+1)\Delta^2(1-d)^2}{4}\Bigg)\Bigg]\\
\end{flalign*}
Let us calculate the cumulative regret of the system from time slot $1$ to $T$. Since we want to get an upper bound of the cumulative regret and check how it scales with time, let us assume we do not accumulate any reward from time slot $t=1$ to $t=\alpha c\ln T$, where, $\alpha > 1$.
\begin{flalign*}
    \mathcal{R}_\mathcal{\epsilon}(T) &=\sum_{t=1}^{T}\mathbb{E}[r^*(t) -r^\mathcal{\epsilon}(t)]\\
    &\leq \sum_{t=1}^{\alpha c\ln T}\Bigg(\frac{p_{k^*}q_{k^*}}{1-d(1-p_{k^*})} - 0\Bigg)\\
    &\quad + \sum_{t=\alpha c\ln T+1}^{T}\Bigg[\frac{p_{k^*}q_{k^*}}{1-d(1-p_{k^*})}\\
    &\quad -p_{k^*}q_{k^*}\Bigg[\frac{1-(d(1-p_{k^*}))^{c\ln T}}{1-d(1-p_{k^*})}\Bigg]\\
    &\quad\times\Bigg(1-3K\frac{c\ln T\ln^2 (t-c\ln T+1)}{t-c\ln T+1}\Bigg)\times\Bigg[1\\
    &\quad -(K-1)\times\exp\Bigg(\frac{-\ln^3 (t-c\ln T+1)\Delta^2(1-d)^2}{4}\Bigg)\Bigg]\Bigg]\\
    &= \frac{p_{k^*}q_{k^*}(\alpha c\ln T)}{1-d(1-p_{k^*})}\\
    &\quad + \frac{p_{k^*}q_{k^*}}{1-d(1-p_{k^*})}\sum_{t=\alpha c\ln T+1}^{T}\Bigg[1- \Big(1-(d(1-p_{k^*}))^{c\ln T}\Big)\\
    &\quad\times\Bigg(1-3K\frac{c\ln T\ln^2 (t-c\ln T+1)}{t-c\ln T+1}\Bigg)\times\Bigg[1\\
    &\quad -(K-1)\times\exp\Bigg(\frac{-\ln^3 (t-c\ln T+1)\Delta^2(1-d)^2}{4}\Bigg)\Bigg]\Bigg]\\
\end{flalign*}
Now, since
\begin{flalign*}
    c &= \max\Bigg\{\frac{-2}{\ln (d(1-p_{k^*})}, \frac{4K}{\Delta^2(1-d)^2}\Bigg\}\\
    \implies c &\geq \frac{-2}{\ln (d(1-p_{k^*}))}\\
    \implies (d(1-p_{k^*}))^{c\ln T}&\leq \frac{1}{T^2}\\
    &\text{(As derived in Theorem \ref{thm: ETC})}
\end{flalign*}
and,
\begin{flalign*}
    c &\geq \frac{4K}{\Delta^2(1-d)^2}\\
    \implies &\frac{c}{K} \geq \frac{4}{\Delta^2(1-d)^2}\\ 
    \exp&\Bigg(\frac{-\ln^3 (t-c\ln T+1)\Delta^2(1-d)^2}{4}\Bigg)\\
    &\leq \exp\Bigg(\frac{-\ln^3 (t-c\ln T+1)}{\frac{c}{K}}\Bigg)
\end{flalign*}
\begin{flalign*}
    \mathcal{R}_\mathcal{\epsilon}(T) &\leq \frac{p_{k^*}q_{k^*}(\alpha c\ln T)}{1-d(1-p_{k^*})}\\
    &\quad + \frac{p_{k^*}q_{k^*}}{1-d(1-p_{k^*})}\sum_{t=\alpha c\ln T+1}^{T}\Bigg[1- \Big(1-\frac{1}{T^2}\Big)\\
    &\quad\times\Bigg(1-3K\frac{c\ln T\ln^2 (t-c\ln T+1)}{t-c\ln T+1}\Bigg)\times\Bigg[1\\
    &\quad-(K-1)\times\exp\Bigg(\frac{-\ln^3 (t-c\ln T+1)}{\frac{c}{K}}\Bigg)\Bigg]\Bigg]\\
    &= \frac{p_{k^*}q_{k^*}}{1-d(1-p_{k^*})}\Bigg[\alpha c\ln T + \sum_{t=\alpha c\ln T+1}^{T}\Bigg[1 -\\
    &\quad\Bigg(1 - (K-1)\exp\Bigg(\frac{-\ln^3 (t-c\ln T+1)}{\frac{c}{K}}\Bigg)\\
    &\qquad - 3K\frac{c\ln T\ln^2 (t-c\ln T+1)}{t-c\ln T+1}\\
    &\quad + 3K\frac{c\ln T\ln^2 (t-c\ln T+1)}{t-c\ln T+1}\\
    &\quad\times(K-1)\exp\Bigg(\frac{-\ln^3 (t-c\ln T+1)}{\frac{c}{K}}\Bigg)\\
    &\quad - \frac{1}{T^2} + \frac{1}{T^2}(K-1)\exp\Bigg(\frac{-\ln^3 (t-c\ln T+1)}{\frac{c}{K}}\Bigg)\\
    &\quad + \frac{1}{T^2}3K\frac{c\ln T\ln^2 (t-c\ln T+1)}{t-c\ln T+1}\\
    &\quad - \frac{1}{T^2}3K\frac{c\ln T\ln^2 (t-c\ln T+1)}{t-c\ln T+1}\\
    &\quad\times(K-1)\exp\Bigg(\frac{-\ln^3 (t-c\ln T+1)}{\frac{c}{K}}\Bigg)\Bigg)\Bigg]\Bigg]\\
    &= \frac{p_{k^*}q_{k^*}}{1-d(1-p_{k^*})}\Bigg[\alpha c\ln T + \sum_{t=\alpha c\ln T+1}^{T}\\
    &\quad\Bigg[(K-1)\exp\Bigg(\frac{-\ln^3 (t-c\ln T+1)}{\frac{c}{K}}\Bigg)\\
    &\qquad + 3K\frac{c\ln T\ln^2 (t-c\ln T+1)}{t-c\ln T+1}\\
    &\quad - 3K\frac{c\ln T\ln^2 (t-c\ln T+1)}{t-c\ln T+1}\\
    &\quad\times(K-1)\exp\Bigg(\frac{-\ln^3 (t-c\ln T+1)}{\frac{c}{K}}\Bigg)\\
    &\quad + \frac{1}{T^2} - \frac{1}{T^2}(K-1)\exp\Bigg(\frac{-\ln^3 (t-c\ln T+1)}{\frac{c}{K}}\Bigg)\\
    &\quad - \frac{1}{T^2}3K\frac{c\ln T\ln^2 (t-c\ln T+1)}{t-c\ln T+1}\\
    &\quad + \frac{1}{T^2}3K\frac{c\ln T\ln^2 (t-c\ln T+1)}{t-c\ln T+1}\\
    &\quad\times(K-1)\exp\Bigg(\frac{-\ln^3 (t-c\ln T+1)}{\frac{c}{K}}\Bigg)\Bigg]\Bigg]\\
    &\leq \frac{p_{k^*}q_{k^*}}{1-d(1-p_{k^*})}\Bigg[\alpha c\ln T + \sum_{t=\alpha c\ln T+1}^{T}\\
    &\quad\Bigg[ (K-1)\exp\Bigg(\frac{-\ln^3 (t-c\ln T+1)}{\frac{c}{K}}\Bigg)\\
    &\qquad + 3K\frac{c\ln T\ln^2 (t-c\ln T+1)}{t-c\ln T+1}\\
    &\quad + \frac{1}{T^2} +\frac{1}{T^2}3K\frac{c\ln T\ln^2 (t-c\ln T+1)}{t-c\ln T+1}\\
    &\quad\times(K-1)\exp\Bigg(\frac{-\ln^3 (t-c\ln T+1)}{\frac{c}{K}}\Bigg)\Bigg]\Bigg]\\
    &= \frac{p_{k^*}q_{k^*}}{1-d(1-p_{k^*})}\Bigg[\alpha c\ln T + \Bigg[(K-1)\\
    &\quad\underbrace{\sum_{t=\alpha c\ln T+1}^{T}\exp\Bigg(\frac{-\ln^3 (t-c\ln T+1)}{\frac{c}{K}}\Bigg)}_\text{$(1)$}\\
    &\qquad + 3Kc\ln T\underbrace{\sum_{t=\alpha c\ln T+1}^{T}\frac{\ln^2 (t-c\ln T+1)}{t-c\ln T+1}}_\text{$(2)$}\\
    &\quad + \underbrace{\sum_{t=\alpha c\ln T+1}^{T}\frac{1}{T^2}}_\text{$(3)$} +\frac{3Kc\ln T (K-1)}{T^2}\sum_{t=\alpha c\ln T+1}^{T}\\
    &\quad\underbrace{\frac{\ln^2 (t-c\ln T+1)}{t-c\ln T+1}\times\exp\Bigg(\frac{-\ln^3 (t-c\ln T+1)}{\frac{c}{K}}\Bigg)}_\text{$(4)$}\Bigg]\Bigg]\\
\end{flalign*}
Let us now calculate the upper bounds on each of the terms individually.

Let us consider $(1)$.
\begin{flalign*}
    \sum_{t=\alpha c\ln T+1}^{T}&\exp\Bigg(\frac{-\ln^3 (t-c\ln T+1)}{\frac{c}{K}}\Bigg)\\
    &\leq \int_{\alpha c\ln T}^{T}\exp\Bigg(\frac{-\ln^3 (t-c\ln T+1)}{\frac{c}{K}}\Bigg)dt\\
    &\leq\int_{\alpha c\ln T}^{T}\exp\Bigg(\frac{-\ln (t-c\ln T+1)}{\frac{c}{K}}\Bigg)dt\\
    &= \frac{c}{K}\int_{\frac{\ln (\alpha c\ln T-c\ln T+1)}{\frac{c}{K}}}^{\frac{\ln(T-c\ln T+1)}{\frac{c}{K}}}\exp\Bigg(-\Big(1-\frac{c}{K}\Big)s\Bigg)ds\\
    &= \frac{\frac{c}{K}}{\Big(1-\frac{c}{K}\Big)}\times\\
    &\quad\Bigg[\exp\Bigg(-\Big(1-\frac{c}{K}\Big)s\Bigg)\Biggr|_{\frac{\ln(T-c\ln T+1)}{\frac{c}{K}}}^{\frac{\ln(\alpha c\ln T-c\ln T+1)}{\frac{c}{K}}}\Bigg]\\
    &= \frac{\frac{c}{K}}{\Big(1-\frac{c}{K}\Big)}\times\\
    &\quad\Bigg[\exp\Bigg(-\Big(1-\frac{c}{K}\Big)\frac{\ln(\alpha c\ln T-c\ln T+1)}{\frac{c}{K}}\Bigg)\\
    &\quad - \exp\Bigg(-\Big(1-\frac{c}{K}\Big)\frac{\ln(T-c\ln T+1)}{\frac{c}{K}}\Bigg)\Bigg]\\
    &= \frac{\frac{c}{K}}{\Big(1-\frac{c}{K}\Big)}\times\\
    &\quad\Bigg[\Bigg(\frac{1}{(\alpha-1)c\ln T+1}\Bigg)^{\frac{1-c/K}{c/K}}\\
    &\quad - \Bigg(\frac{1}{T-c\ln T+1}\Bigg)^{\frac{1-c/K}{c/K}}\Bigg]
\end{flalign*}
Let us consider $(2)$.
\begin{flalign*}
    \sum_{t=\alpha c\ln T+1}^{T}\frac{\ln^2 (t-c\ln T+1)}{t-c\ln T+1} &\leq \int_{\alpha c\ln T}^{T}\frac{\ln^2 (t-c\ln T+1)}{t-c\ln T+1}dt\\
    &= \int_{\ln(\alpha c\ln T-c\ln T+1)}^{\ln(T-c\ln T+1)}s^2ds\\
    &= \frac{1}{3}\Big(\ln^3(T-c\ln T+1)\\
    &\quad - \ln^3(\alpha c\ln T-c\ln T+1)\Big)
\end{flalign*}
Let us consider $(3)$.
\begin{flalign*}
    \sum_{t=\alpha c\ln T+1}^{T}\frac{1}{T^2} &= \frac{1}{T^2}(T-\alpha c\ln T)\\
    &= \frac{1}{T} - \alpha c\frac{\ln T}{T^2}
\end{flalign*}
Let us consider $(4)$.
\begin{flalign*}
    &\sum_{t=\alpha c\ln T+1}^{T}\frac{\ln^2 (t-c\ln T+1)}{t-c\ln T+1}\exp\Bigg(\frac{-\ln^3 (t-c\ln T+1)}{\frac{c}{K}}\Bigg)\\
    &\leq \int_{t=\alpha c\ln T}^{T}\frac{\ln^2 (t-c\ln T+1)}{t-c\ln T+1}\exp\Bigg(\frac{-\ln^3 (t-c\ln T+1)}{\frac{c}{K}}\Bigg)dt\\
    &= \frac{c}{3K}\int_{\frac{\ln^3(\alpha c\ln T-c\ln T+1)}{\frac{c}{K}}}^{\frac{\ln^3(T-c\ln T+1)}{\frac{c}{K}}}\exp(-s)ds\\
    &= \frac{c}{3K}\Bigg[\exp(-s)\Biggr|_{\frac{\ln^3(T-c\ln T+1)}{\frac{c}{K}}}^{\frac{\ln^3(\alpha c\ln T-c\ln T+1)}{\frac{c}{K}}}\Bigg]\\
    &= \frac{c}{3K}\Bigg[\exp\Bigg({\frac{-\ln^3(\alpha c\ln T-c\ln T+1)}{\frac{c}{K}}}\Bigg)\\
    &\quad -\exp\Bigg({\frac{-\ln^3(T-c\ln T+1)}{\frac{c}{K}}}\Bigg)\Bigg]
\end{flalign*}
Substituting the above $4$ expressions in the regret equation, we get
\begin{flalign*}
    \mathcal{R}_\epsilon(T) &\leq \frac{p_{k^*}q_{k^*}}{1-d(1-p_{k^*})}\Bigg[\alpha c\ln T + (K-1)\\
    &\quad\times\frac{\frac{c}{K}}{\Big(1-\frac{c}{K}\Big)}\Bigg[\Bigg(\frac{1}{(\alpha-1)c\ln T+1}\Bigg)^{\frac{1-c/K}{c/K}}\\
    &\quad - \Bigg(\frac{1}{T-c\ln T+1}\Bigg)^{\frac{1-c/K}{c/K}}\Bigg]\\
    &\quad + Kc\ln T\Big(\ln^3(T-c\ln T+1)\\
    &\quad - \ln^3(\alpha c\ln T-c\ln T+1)\Big)\\
    &\quad + \frac{1}{T} - \alpha c\frac{\ln T}{T^2} +\frac{c^2\ln T (K-1)}{T^2}\\
    &\quad\times\Bigg[\exp\Bigg({\frac{-\ln^3(\alpha c\ln T-c\ln T+1)}{\frac{c}{K}}}\Bigg)\\
    &\quad -\exp\Bigg({\frac{-\ln^3(T-c\ln T+1)}{\frac{c}{K}}}\Bigg)\Bigg]\Bigg]\\
    &\leq \frac{p_{k^*}q_{k^*}}{1-d(1-p_{k^*})}\Bigg[\alpha c\ln T + (K-1)\\
    &\quad\times\frac{\frac{c}{K}}{\Big(1-\frac{c}{K}\Big)}\Bigg(\frac{1}{(\alpha-1)c\ln T+1}\Bigg)^{\frac{1-c/K}{c/K}}\\
    &\quad + Kc\ln T\ln^3(T-c\ln T+1)\\
    &\quad + \frac{1}{T} +\frac{c^2\ln T (K-1)}{T^2}\\
    &\quad\times\exp\Bigg({\frac{-\ln^3(\alpha c\ln T-c\ln T+1)}{\frac{c}{K}}}\Bigg)\Bigg]\\
\end{flalign*}
\end{proof}

\subsection{Proof of Theorem \ref{thm:lower bound}}
\textbf{Theorem 4} (Lower bound on any $\gamma$- consistent policy). Given a problem instance $(p_k,q_k,d,K)$, let $\mu_k = \frac{p_kq_k}{1-d(1-p_k)} ~ \forall ~ 1 \leq k \leq K$, $\mu_{k_{\text{min}}} = \min\limits_{k=1:K}\mu_k$, $k^*=\arg \max\limits_{1 \leq k \leq K}\mu_k$ (as defined in Theorem \ref{thm1}), $\mu_{k^*} = \max\limits_{k=1:K}\mu_k = \frac{p_{k^*}q_{k^*}}{1-d(1-p_{k^*})}$, $\Delta = \mu_{k^*} - \max\limits_{k\neq k^*} \mu_k$, $\Delta_p = \min\limits_{k\neq k^*}p_k-p_{k^*}$, $p_\text{min} = \min\limits_{k=1:K}p_k$, $q_{\text{min}} = \min\limits_{k=1:K}q_k$. For any $\gamma$- consistent policy $\mathcal{P}$,
\begin{flalign*}
    \mathcal{R}_\mathcal{P}(T)&\geq \frac{(K-1)D(\mu)\Delta_p}{\Delta}\mu_{k^*}\\
    &\quad\times d((1-\gamma)\log T - \log (4KC))\\
    &\quad +(p_{k^*}q_{k^*} -p_{\text{min}}q_{\text{min}})T
\end{flalign*}
\hspace*{4cm} where, $D(\mu) = \frac{\Delta}{KL\Big(\mu_{k_{\text{min}}},\frac{\mu_{k^*}+1}{2}\Big)}$
\begin{proof}
    To prove this theorem, we construct an alternative service process for our system model such that under any scheduling policy, the reward evolution for the system with this alternative service process has the same distribution as that for the original system.\\
    The alternative service process is constructed as follows:\\
    Let $\{U(t)\}_{t \geq 1}$ and $\{V(t)\}_{t \geq 1}$ be i.i.d. random variables distributed uniformly in [0,1]. Let us assume we schedule source $k$ for transmission in time slot $t$. Let us denote the following services in the alternative process for source $k$ at every time slot $t$:
    \begin{enumerate}
        \item [(i)] The successful transmission of the update by source $k$ in time slot $t$ is given by $Y_k(t) = \mathds{1}_{\{U(t) \leq p_k\}}$ for all $t$.
        \item [(ii)] The correctness of the measured update by source $k$ at time slot $t$ is given by $Z_k(t) = \mathds{1}_{\{V(t) \leq q_k\}}$ for all $t$.
    \end{enumerate}
    Note that under this alternative service process, $\mathbb{E}[Y_k(t)] = p_k$ and $\mathbb{E}[Z_k(t)] = q_k$. Hence the marginals of the service offered by source $k$ under this alternative process is the same as that of the original system.\\
    Now, let us assume that the reward in time slot $t$ under any $\gamma$-consistent policy, $\mathcal{P}$, and the oracle policy be denoted by $r^\mathcal{P}(t)$ and $r^*(t)$ respectively. Further, let $M(t)$ and $M^*(t)$ be indicator random variables denoting successful transmission of update in time slot $t$ by a $\gamma$-consistent policy and the oracle policy respectively. Also, let $N(t)$ and $N^*(t)$ be indicator random variables denoting correct measurement of update in time slot $t$ by a $\gamma$-consistent policy and the oracle policy respectively.\\
    Hence,
    \begin{flalign*}
        r^*(t) &= (1-M^*(t))r^*(t-1)d + M^*(t)N^*(t)\\
        r^\mathcal{P}(t) &= (1-M(t))r(t-1)d + M(t)N(t)
    \end{flalign*}
    \begin{flalign*}
    \implies r^*(t)-r^\mathcal{P}(t) &= [(1-M^*(t))r^*(t-1)d+M^*(t)N^*(t)]\\
    &\quad - [(1-M(t))r(t-1)d+M(t)N(t)]\\
    &= (1-M^*(t))r^*(t-1)d-(1-M(t))\\
    &\quad \times r(t-1)d +M^*(t)N^*(t)-M(t)N(t)
\end{flalign*}
In the coupled system, $r^*(t) \geq r^\mathcal{P}(t)~ \forall~ t$, hence,
\begin{flalign*}
    r^*(t)-r^\mathcal{P}(t) &\geq (1-M^*(t))r^*(t-1)d-(1-M(t))r^*(t-1)d\\
    &\quad +M^*(t)N^*(t)-M(t)N(t)\\
    &= (M(t)-M^*(t))r^*(t-1)d+M^*(t)N^*(t)\\
    &\qquad -M(t)N(t)
\end{flalign*}
Taking expectations on both sides, we get
\begin{flalign*}
    \mathbb{E}[r^*(t)-r^\mathcal{P}(t)] &\geq \mathbb{E}[(M(t)-M^*(t))r^*(t-1)d\\
    &\quad +M^*(t)N^*(t) -M(t)N(t)]\\
    &= \mathbb{E}[(M(t)-M^*(t))r^*(t-1)d]\\
    &\quad +\mathbb{E}[M^*(t)N^*(t)]-\mathbb{E}[M(t)N(t)]
\end{flalign*}
Since the reward received in time slot $t-1$ under the Oracle policy is independent of the successful transmission of updates at time slot $t$ under the $\gamma$- consistent and Oracle policy, we get
\begin{flalign*}
    \mathbb{E}[r^*(t)-r^\mathcal{P}(t)] &\geq \mathbb{E}[M(t)-M^*(t)]\mathbb{E}[r^*(t-1)d]\\
    &\quad +\mathbb{E}[M^*(t)N^*(t)]-\mathbb{E}[M(t)N(t)]\\
    &= d~\mathbb{E}[M(t)-M^*(t)]\mathbb{E}[r^*(t-1)]\\
    &\quad +\mathbb{E}[M^*(t)N^*(t)]-\mathbb{E}[M(t)N(t)]
\end{flalign*}
Since the expected value of reward received under the oracle policy at all times is $\mathbb{E}[r^*(t)] = \frac{p_{k^*}q_{k^*}}{1-d(1-p_{k^*})} ~ \forall ~ t$
\begin{flalign*}
    \mathbb{E}[r^*(t)-r^\mathcal{P}(t)] &\geq \frac{p_{k^*}q_{k^*}d}{1-d(1-p_{k^*})}\mathbb{E}[M(t)-M^*(t)]\\
    &\quad +\mathbb{E}[M^*(t)N^*(t)]-\mathbb{E}[M(t)N(t)]
\end{flalign*}
Now, taking summation on both sides from $t=1$ to $t=T$, we get
\begin{flalign}
    \sum_{t=1}^{T}\mathbb{E}[r^*(t)-r^\mathcal{P}(t)] &\geq \sum_{t=1}^{T}\Bigg(\frac{p_{k^*}q_{k^*}d}{1-d(1-p_{k^*})}\mathbb{E}[M(t)-M^*(t)]\nonumber\\
    &\quad +\mathbb{E}[M^*(t)N^*(t)]-\mathbb{E}[M(t)N(t)]\Bigg)\nonumber\\
    \mathcal{R}_\mathcal{P}(T) &\geq \frac{p_{k^*}q_{k^*}d}{1-d(1-p_{k^*})}\sum_{t=1}^{T}\mathbb{E}[M(t)-M^*(t)]\nonumber\\
    &\quad +\sum_{t=1}^{T}\mathbb{E}[M^*(t)N^*(t)]\nonumber\\
    &\quad-\sum_{t=1}^{T}\mathbb{E}[M(t)N(t)] \label{eq:lower bound}
\end{flalign}
Let $P_k(t)$ be an indicator random variable denoting if the update sent by source $k$ in time slot $t$ is successful and let $P_{k^*}(t)$ be an indicator random variable denoting if the update sent by the optimal source, $k^*$, in time slot $t$ is successful. On similar lines, let us define $Q_k(t)$ and $Q_{k^*}(t)$ to be indicator random variables denoting if the update measured by the source $k$ and the optimal source $k^*$ in time slot $t$ is correct respectively. Also, let us assume $k(t)$ to be the index of the source scheduled by the $\gamma$- consistent policy in time slot $t$. These definitions follow some basic relations as given:
\begin{flalign*}
    M(t) &= P_{k(t)}(t)\\
    &= \sum_{k=1}^{K}\mathds{1}_{\{k(t)=k\}}P_k(t)\\
    M^*(t) &= P_{k^*}(t)\\
    N(t) &= Q_{k(t)}(t)\\
    &=\sum_{k=1}^{K}\mathds{1}_{\{k(t)=k\}}Q_k(t)\\
    N^*(t) &= Q_{k^*}(t)
\end{flalign*}
Substituting these expressions in \ref{eq:lower bound}, we get
\begin{flalign*}
    \mathcal{R}_\mathcal{P}(T) &\geq \frac{p_{k^*}q_{k^*}d}{1-d(1-p_{k^*})}\sum_{t=1}^{T}\mathbb{E}\Bigg[\sum_{k=1}^{K}\mathds{1}_{\{k(t)=k\}}P_k(t)-P_{k^*}(t)\Bigg]\\
    &\quad +\sum_{t=1}^{T}\mathbb{E}[P_{k^*}(t)Q_{k^*}(t)]\\
    &\quad -\sum_{t=1}^{T}\mathbb{E}\Bigg[\sum_{k=1}^{K}\mathds{1}_{\{k(t)=k\}}P_k(t)\sum_{k=1}^{K}\mathds{1}_{\{k(t)=k\}}Q_k(t)\Bigg]\\
    &= \frac{p_{k^*}q_{k^*}d}{1-d(1-p_{k^*})}\underbrace{\sum_{t=1}^{T}\mathbb{E}\Bigg[\sum_{k=1}^{K}\mathds{1}_{\{k(t)=k\}}P_k(t)-P_{k^*}(t)\Bigg]}_\text{$(1)$}\\
    &\quad +\underbrace{\sum_{t=1}^{T}\mathbb{E}[P_{k^*}(t)Q_{k^*}(t)]}_\text{$(2)$}\\
    &\quad -\underbrace{\sum_{t=1}^{T}\mathbb{E}\Bigg[\sum_{k=1}^{K}\mathds{1}_{\{k(t)=k\}}\Big(P_k(t)Q_k(t)\Big)\Bigg]}_\text{$(3)$}
\end{flalign*}
Let us now consider each of the terms individually.\\
Let us consider $(1)$.
\begin{flalign*}
    \sum_{t=1}^{T}\mathbb{E}&\Bigg[\sum_{k=1}^{K}\mathds{1}_{\{k(t)=k\}}P_k(t)-P_{k^*}(t)\Bigg]\\
    &= \sum_{t=1}^{T}\mathbb{E}\Bigg[\sum_{\substack{k=1 \\~ k\neq k^*}}^{K}\mathds{1}_{\{k(t)=k\}}(P_k(t)-P_{k^*}(t))\Bigg]\\
    &= \sum_{t=1}^{T}\Bigg[\sum_{\substack{k=1 \\~ k\neq k^*}}^{K}\mathbb{E}\Big[\mathds{1}_{\{k(t)=k\}}(P_k(t)-P_{k^*}(t))\Big]\Bigg]\\
    &= \sum_{t=1}^{T}\Bigg[\sum_{\substack{k=1 \\~ k\neq k^*}}^{K}\mathbb{E}\Big[\mathbb{E}\Big[\mathds{1}_{\{k(t)=k\}}(P_k(t)\\
    &\qquad\qquad\qquad\qquad\qquad -P_{k^*}(t))|\mathds{1}_{\{k(t)=k\}}\Big]\Big]\Bigg]\\
    &~~~~~~~~~~~~~~~~~~~~~~~~~~~~~(\text{Law of Iterated Expectations})\\
    &= \sum_{t=1}^{T}\Bigg[\sum_{\substack{k=1 \\~ k\neq k^*}}^{K}\mathbb{E}\Big[\mathds{1}_{\{k(t)=k\}}\mathbb{E}\Big[(P_k(t)\\
    &\qquad\qquad\qquad\qquad\qquad -P_{k^*}(t))|\mathds{1}_{\{k(t)=k\}}\Big]\Big]\Bigg]\\
    &= \sum_{t=1}^{T}\Bigg[\sum_{\substack{k=1 \\~ k\neq k^*}}^{K}\mathbb{E}\Big[\mathds{1}_{\{k(t)=k\}}\mathbb{E}\Big[(P_k(t)|\mathds{1}_{\{k(t)=k\}}\\
    &\qquad\qquad\qquad\qquad\qquad\qquad -P_{k^*}(t)|\mathds{1}_{\{k(t)=k\}})\Big]\Big]\Bigg]\\
    &= \sum_{t=1}^{T}\Bigg[\sum_{\substack{k=1 \\~ k\neq k^*}}^{K}\mathbb{E}\Big[\mathds{1}_{\{k(t)=k\}}\Big(\mathbb{E}\Big[P_k(t)|\mathds{1}_{\{k(t)=k\}}\Big]\\
    &\qquad\qquad\qquad\qquad\qquad -\mathbb{E}\Big[P_{k^*}(t)|\mathds{1}_{\{k(t)=k\}}\Big]\Big)\Big]\Bigg]\\
    &= \sum_{t=1}^{T}\Bigg[\sum_{\substack{k=1 \\~ k\neq k^*}}^{K}\mathbb{E}\Big[\mathds{1}_{\{k(t)=k\}}\Big(\mathbb{E}\Big[\mathds{1}_{\{U(t)\leq p_k\}}|\mathds{1}_{\{k(t)=k\}}\Big]\\
    &\qquad\qquad\qquad\qquad\qquad -\mathbb{E}\Big[\mathds{1}_{\{U(t)\leq p_{k^*}\}}|\mathds{1}_{\{k(t)=k\}}\Big]\Big)\Big]\Bigg]\\
    &= \sum_{t=1}^{T}\Bigg[\sum_{\substack{k=1 \\~ k\neq k^*}}^{K}\mathbb{E}\Big[\mathds{1}_{\{k(t)=k\}}(p_k-p_{k^*})\Big]\Bigg]\\
    &= \sum_{t=1}^{T}\Bigg[\sum_{\substack{k=1 \\~ k\neq k^*}}^{K}(p_k-p_{k^*})\mathbb{E}\Big[\mathds{1}_{\{k(t)=k\}}\Big]\Bigg]\\
    &\geq \sum_{t=1}^{T}\Bigg[\sum_{\substack{k=1 \\~ k\neq k^*}}^{K}\Big(\min\limits_{k\neq k^*}p_k-p_{k^*}\Big)\mathbb{E}\Big[\mathds{1}_{\{k(t)=k\}}\Big]\Bigg]\\
    &= \sum_{t=1}^{T}\Bigg[\sum_{\substack{k=1 \\~ k\neq k^*}}^{K}\Delta_p\mathbb{E}\Big[\mathds{1}_{\{k(t)=k\}}\Big]\Bigg]\\
    &~~~~~~~~~~~~~~~~~~~~~~~\text{where, $\Delta_p = \min\limits_{k\neq k^*}p_k-p_{k^*}$}\\
    &= \Delta_p\sum_{\substack{k=1 \\~ k\neq k^*}}^{K}\sum_{t=1}^{T}\mathbb{E}\Big[\mathds{1}_{\{k(t)=k\}}\Big]
\end{flalign*}
Let us consider $(2)$.
\begin{flalign*}
    \sum_{t=1}^{T}\mathbb{E}[P_{k^*}(t)Q_{k^*}(t)] &= \sum_{t=1}^{T}\mathbb{E}[P_{k^*}(t)]\mathbb{E}[Q_{k^*}(t)]\\
    &= \sum_{t=1}^{T}p_{k^*}q_{k^*}\\
    &= p_{k^*}q_{k^*}\sum_{t=1}^{T}1\\
    &= p_{k^*}q_{k^*}T
\end{flalign*}
Let us consider $(3)$.
\begin{flalign*}
    \sum_{t=1}^{T}&\mathbb{E}\Bigg[\sum_{k=1}^{K}\mathds{1}_{\{k(t)=k\}}\Big(P_k(t)Q_k(t)\Big)\Bigg]\\
    &= \sum_{t=1}^{T}\sum_{k=1}^{K}\mathbb{E}\Big[\mathds{1}_{\{k(t)=k\}}\Big(P_k(t)Q_k(t)\Big)\Big]\\
    &= \sum_{t=1}^{T}\sum_{k=1}^{K}\mathbb{E}\Bigg[\mathbb{E}\Big[\mathds{1}_{\{k(t)=k\}}\Big(P_k(t)Q_k(t)\Big)\Big|\mathds{1}_{\{k(t)=k\}}\Big]\Bigg]\\
    &= \sum_{t=1}^{T}\sum_{k=1}^{K}\mathbb{E}\Bigg[\mathds{1}_{\{k(t)=k\}}\mathbb{E}\Big[P_k(t)Q_k(t)\Big|\mathds{1}_{\{k(t)=k\}}\Big]\Bigg]\\
    &= \sum_{t=1}^{T}\sum_{k=1}^{K}\mathbb{E}\Bigg[\mathds{1}_{\{k(t)=k\}}\mathbb{E}\Big[\mathds{1}_{\{U(t)\leq p_k\}}\mathds{1}_{\{V(t)\leq q_k\}}\Big|\mathds{1}_{\{k(t)=k\}}\Big]\Bigg]\\
    &= \sum_{t=1}^{T}\sum_{k=1}^{K}\mathbb{E}\Bigg[\mathds{1}_{\{k(t)=k\}}p_kq_k\Bigg]\\
    &\geq \sum_{t=1}^{T}\sum_{k=1}^{K}\mathbb{E}\Bigg[\mathds{1}_{\{k(t)=k\}}p_{\text{min}}q_{\text{min}}\Bigg]\\
    &~~~~~~~~~~~~~~~~~~~~~~\text{where, $p_{\text{min}} = \min\limits_{k=1:K}p_k$, $q_{\text{min}} = \min\limits_{k=1:K}q_k$}\\
    &= p_{\text{min}}q_{\text{min}}\sum_{t=1}^{T}\sum_{k=1}^{K}\mathbb{E}\Big[\mathds{1}_{\{k(t)=k\}}\Big]\\
    &= p_{\text{min}}q_{\text{min}}\sum_{t=1}^{T}\sum_{k=1}^{K}\mathbb{P}(k(t)=k)\\
    &= p_{\text{min}}q_{\text{min}}\sum_{t=1}^{T}1\\
    &= p_{\text{min}}q_{\text{min}}T
\end{flalign*}
Substituting the above $3$ expressions in the regret equation, we get
\begin{flalign*}
    \mathcal{R}_\mathcal{P}(T)&\geq \frac{p_{k^*}q_{k^*}d}{1-d(1-p_{k^*})}\Delta_p\sum_{\substack{k=1 \\~ k\neq k^*}}^{K}\sum_{t=1}^{T}\mathbb{E}\Big[\mathds{1}_{\{k(t)=k\}}\Big]\\
    &\quad +p_{k^*}q_{k^*}T -p_{\text{min}}q_{\text{min}}T\\
    &= \frac{p_{k^*}q_{k^*}d}{1-d(1-p_{k^*})}\Delta_p\sum_{\substack{k=1 \\~ k\neq k^*}}^{K}\mathbb{E}[T_k(T+1)]\\
    &\quad +(p_{k^*}q_{k^*} -p_{\text{min}}q_{\text{min}})T\\
    &\text{where, $T_k(T+1)$ is the number of time slots in which}\\
    &\text{source $k$ has been scheduled in the time interval $1$ to $T$.}
\end{flalign*}
We use the following result from \cite{fatale2021regret} for calculating the expression of $\mathbb{E}[T_k(T+1)]$.
\begin{lemma}
    Let $T_k(T)$ be the number of time slots in which source $k$ is scheduled in the time interval $1$ to $T-1$. For a problem instance $\mu (=\mu_k ~\forall ~k \in [K])$, let $\mu_{k_{\text{min}}} = \min\limits_{k=1:K}\mu_k > 0$ and $\mu_{k^*} = \max\limits_{k=1:K}\mu_k$. For any $\gamma$-consistent policy $\mathcal{P}$, there exist constant $\tau$ and $C$, s.t. for any $t > \tau$,
    \begin{equation*}
        \Delta\sum_{\substack{k=1 \\~ k\neq k^*}}^{K}\mathbb{E}[T_k(T+1)] \geq (K-1)D(\mu)((1-\gamma)\log t - \log (4KC))
    \end{equation*}
    where $D(\mu) = \frac{\Delta}{KL\Big(\mu_{k_{\text{min}}}, \frac{\mu_{k^*}+1}{2}\Big)}$ and $\Delta = \mu_{k^*} - \max\limits_{k\neq K^*}\mu_k$.
\end{lemma}
Hence, we get
\begin{flalign*}
    \mathcal{R}_\mathcal{P}(T)&\geq\frac{p_{k^*}q_{k^*}d}{1-d(1-p_{k^*})}\Delta_p\sum_{\substack{k=1 \\~ k\neq k^*}}^{K}\mathbb{E}[T_k(T+1)]\\
    &\quad +(p_{k^*}q_{k^*} -p_{\text{min}}q_{\text{min}})T\\
    &\geq \frac{p_{k^*}q_{k^*}d}{1-d(1-p_{k^*})}\frac{\Delta_p}{\Delta}(K-1)D(\mu)\\
    &\quad\times((1-\gamma)\log T - \log (4KC))\\
    &\quad +(p_{k^*}q_{k^*} -p_{\text{min}}q_{\text{min}})T\\
    &= \frac{(K-1)D(\mu)\Delta_p}{\Delta}\mu_{k^*}\\
    &\quad\times d((1-\gamma)\log T - \log (4KC))\\
    &\quad +(p_{k^*}q_{k^*} -p_{\text{min}}q_{\text{min}})T\\
    &~~~~~~~~~~~~~~~~~~~\text{where,}~ D(\mu) = \frac{\Delta}{KL\Big(\mu_{k_{\text{min}}},\frac{\mu_{k^*}+1}{2}\Big)}
\end{flalign*}
\end{proof}

\end{document}